%% file: main.tex
\documentclass[nohyperref]{article}

\usepackage{microtype}
\usepackage{graphicx}
\usepackage{subfigure}
\usepackage{booktabs} 

\usepackage{hyperref}

\usepackage[accepted]{icml2022}

\usepackage{amsmath}
\usepackage{amssymb}
\usepackage{mathtools}
\usepackage{amsthm}
\usepackage{pifont}
\usepackage{thmtools,thm-restate}
\usepackage{multirow}
\usepackage{multicol}
\usepackage{enumitem}
\usepackage{times}
\usepackage{bbm,bm}
\usepackage{soul}
\usepackage{url}
\usepackage[small]{caption}
\usepackage{float}
\usepackage{tabularx}
\usepackage{pifont}
\usepackage{algorithm}
\usepackage{algorithmic}

\usepackage[capitalize,noabbrev]{cleveref}

\theoremstyle{plain}
\newtheorem{theorem}{Theorem}[section]
\newtheorem{proposition}[theorem]{Proposition}

\theoremstyle{definition}
\newtheorem{definition}[theorem]{Definition}

\theoremstyle{remark}

\usepackage[textsize=tiny]{todonotes}

\input{math_commands}

\input{commands}

\newcommand{\cmark}{\ding{51}}%
\newcommand{\xmark}{\ding{55}}%

\icmltitlerunning{Plan Your Target and Learn Your Skills: Transferable State-Only Imitation Learning via Decoupled Policy Optimization}

\begin{document}

\twocolumn[
\icmltitle{Plan Your Target and Learn Your Skills: Transferable State-Only \\ Imitation Learning via Decoupled Policy Optimization}




\begin{icmlauthorlist}
\icmlauthor{Minghuan Liu}{yyy}
\icmlauthor{Zhengbang Zhu}{yyy}
\icmlauthor{Yuzheng Zhuang}{comp}
\icmlauthor{Weinan Zhang}{yyy}
\icmlauthor{Jianye Hao}{comp,sch}
\icmlauthor{Yong Yu}{yyy}
\icmlauthor{Jun Wang}{comp}
\end{icmlauthorlist}

\icmlaffiliation{yyy}{Shanghai Jiao Tong University}
\icmlaffiliation{comp}{Huawei Noah's Ark Lab}
\icmlaffiliation{sch}{Tianjin University}

\icmlcorrespondingauthor{Weinan Zhang}{wnzhang@sjtu.edu.cn}

\icmlkeywords{Machine Learning, ICML}

\vskip 0.3in
]



\printAffiliationsAndNotice{}  

\input{papers/0-abs}
\input{papers/1-intro}
\input{papers/2-related}
\input{papers/3-preliminary}
\input{papers/4-method}
\input{papers/5-exp}
\input{papers/6-conclusion}

\bibliographystyle{icml2022}
\bibliography{ref}

\newpage
\appendix
\onecolumn
\input{papers/7-appendix}
\end{document}

%% file: math_commands.tex
\usepackage{amsmath,amsfonts,bm}

\def\eqref#1{equation~\ref{#1}}

\def\1{\bm{1}}

\DeclareMathAlphabet{\mathsfit}{\encodingdefault}{\sfdefault}{m}{sl}
\SetMathAlphabet{\mathsfit}{bold}{\encodingdefault}{\sfdefault}{bx}{n}

\DeclareMathOperator*{\argmin}{arg\,min}

%% file: commands.tex
\newcommand{\fig}[1]{Fig.~\ref{#1}}
\newcommand{\eq}[1]{Eq.~(\ref{#1})}
\newcommand{\tb}[1]{Tab.~\ref{#1}}
\newcommand{\se}[1]{Section~\ref{#1}}
\newcommand{\ap}[1]{Appendix~\ref{#1}}

\newcommand{\prop}[1]{Proposition~\ref{#1}}

\newcommand{\theo}[1]{Theorem~\ref{#1}}

\newcommand*{\dif}{\mathop{}\!\mathrm{d}}

\newcommand{\bbE}{\ensuremath{\mathbb{E}}} 
\newcommand{\caA}{\ensuremath{\mathcal{A}}} 
\newcommand{\caS}{\ensuremath{\mathcal{S}}} 
\newcommand{\caM}{\ensuremath{\mathcal{M}}} 
\newcommand{\caD}{\ensuremath{\mathcal{D}}} 
\newcommand{\caL}{\ensuremath{\mathcal{L}}} 
\newcommand{\caT}{\ensuremath{\mathcal{T}}} 
\newcommand{\caB}{\ensuremath{\mathcal{B}}} 
\newcommand{\kld}{\text{D}_{\text{KL}}} 
\newcommand{\jsd}{\text{D}_{\text{JS}}}
\newcommand{\fd}{\text{D}_{\text{f}}} 

\newcommand{\piE}{{\pi_E}}

%% file: papers/0-abs.tex
\begin{abstract}
Recent progress in state-only imitation learning extends the scope of applicability of imitation learning to real-world settings by relieving the need for observing expert actions.
However, existing solutions only learn to extract a state-to-action mapping policy from the data, without considering how the expert plans to the target. This hinders the ability to leverage demonstrations and limits the flexibility of the policy.
In this paper, we introduce Decoupled Policy Optimization (DePO), which explicitly decouples the policy as a high-level state planner and an inverse dynamics model. With embedded decoupled policy gradient and generative adversarial training, DePO enables knowledge transfer to different action spaces or state transition dynamics, and can generalize the planner to out-of-demonstration state regions.
Our in-depth experimental analysis shows the effectiveness of DePO on learning a generalized target state planner while achieving the best imitation performance. We demonstrate the appealing usage of DePO for transferring across different tasks by pre-training, and the potential for co-training agents with various skills.


\end{abstract}

%% file: papers/1-intro.tex
\section{Introduction}
\label{sec:intro}


Imitation Learning (IL) offers a way to train an agent from demonstrations by mimicking the expert's behaviors without constructing hand-crafted reward functions~\cite{hussein2017imitation,liu2020energy}, which requires the expert demonstrations to include information of not only states but also actions. Unfortunately, the action information is absent from many real-world demonstration resources, e.g., traffic surveillance and sport records. To tackle this challenge, one of the potential solutions is state-only imitation learning (SOIL), also known as learning from observations (LfO)~\cite{torabi2019recent}, which extends the scope of applicability of IL by relieving the memorization of low-level actions.
However, existing SOIL methods~\citep{torabi2018behavioral, yang2019imitation} only attempt to match the expert state sequences implicitly by determining feasible actions for each pair of consecutive states, which is an ad-hoc solution for a specific task and ignores understanding the high-level target planning strategy of the expert. 



Evidence from humans' cognition and learning in the physical world suggests that human intelligence is effective at transferring high-level knowledge via observations by extracting invariant characteristics across tasks and performing long-term planning that requires different skills. For example, a courier can learn from peers about the optimal delivery path while inheriting his own skills and habits for driving an electric vehicle or riding a bike. Inspired by this insight, to model a generalized target planner and allow transferring to various action spaces and dynamics, we propose Decoupled Policy Optimization (DePO), a novel architecture that decouples a state-to-action policy as two modules -- a state planner that generates the consecutive target state, followed by an inverse dynamics model that delivers action to achieve the target state.
Intuitively, the state planner is prompted to perform action-aware planning with respect to the inverse dynamics modelling.
Furthermore, to prevent the state planner from compounding error by vanilla supervised learning the state transitions from demonstrations, DePO combines the two modules as an integral policy function and incorporates generative adversarial trainining with the induced decoupled policy gradient. 

As such, DePO provides generalized plannable state predictions on out-of-demonstration state regions, guiding the agent to match the expert state sequences.
The flexibility of DePO supports transferring the planner over homogeneous tasks with different skills.
In experiments, we conduct in-depth analysis showing that DePO enjoys several advantages:
\begin{itemize}
\vspace{-10pt}
    \item DePO has a generalized ability of state planning on out-of-demonstration states.
    \vspace{-5pt}
    \item DePO learns accurate state predictions while keeping the best imitation performances.
    \vspace{-5pt}
    \item DePO allows pre-training and co-training with less sampling cost for agents over various action spaces and dynamics.
\end{itemize}

%% file: papers/2-related.tex
\section{Related Work}
\label{sec:related}




SOIL endows the agent with the ability to learn from expert states. Although lacking expert decision information, most of previous works still optimize a state-to-action mapping policy to match the expert state transition occupancy measure (OM)~\citep{ho2016generative}.
\citet{torabi2018behavioral} trained an inverse model to label the action information and applied behavioral cloning, while \citet{torabi2019adversarial} generalized GAIL to match the state transition OM.
\citet{yang2019imitation} analyzed the inverse dynamics mismatch in SOIL and introduced mutual information to narrow it. 
\citet{huang2019learning} applied SOIL on autonomous driving tasks by utilizing a hierarchical policy with a neural decision module and a non-differentiable execution module.

\begin{table}[h!]
\vspace{-2pt}
\caption{Comparison of different methods.}
\vspace{-4pt}
\label{tb:inverse-dynamics-comparison}
\centering
\resizebox{0.48\textwidth}{!}{
\begin{tabular}{ccccc}
\toprule
\multirow{2}{*}{Method} & Inverse & State & Decoupled & Main \\
& Dynamics & Planner & Policy & Task\\
\midrule
BCO \citep{torabi2018behavioral} & \cmark & \xmark & \xmark & SOIL \\
GAIfO \citep{torabi2019adversarial} & \xmark & \xmark & \xmark & SOIL \\
IDDM \citep{yang2019imitation} & \xmark & \xmark & \xmark & SOIL \\
OPOLO \citep{zhu2020off} & \cmark & \xmark & \xmark & SOIL \\
PID-GAIL \citep{huang2019learning} & \xmark & \xmark & \cmark & IL \\
QSS \citep{edwards2020estimating} & \cmark & \cmark & \cmark & RL \\
SAIL \citep{liu2019state} & \cmark & \cmark & \xmark & IL \\
GSP \citep{pathak2018zero} & \cmark & \xmark & \xmark & SOIL \\
IMO \citep{kimura2018internal} & \xmark & \cmark & \xmark & SOIL \\
\midrule
DePO (Ours) & \cmark & \cmark & \cmark & SOIL \\
\bottomrule
\end{tabular}
}
\vspace{-6pt}
\end{table}

Our work decouples the state-to-action policy into two modules, i.e., the inverse dynamics model and the state transition planner. Both modules have been widely used by many previous works on RL and IL tasks. For instance, \citet{torabi2018behavioral} and \citet{guo2019hybrid} proposed Behavioral Cloning from Observations (BCO), which trained an inverse dynamics model to label the state-only demonstrations with inferred actions.
\citet{nair2017combining} proposed to match a human-specified image sequence of ropes manipulating with an inverse dynamics model.
\citet{pathak2018zero} proposed GSP, which trained a multi-step inverse dynamics model from exploration which is regularized by cycle consistency for image-based imitation. From the task perspective, they try to exactly match an expert demonstration sequence, with a binary classifier judging whether each goal is reached.
\citet{kimura2018internal} utilized a state transition predictor to fit the state transition probability in the expert data, which is further used to compute a predefined reward function.
\citet{liu2019state} constructed a policy prior using the inverse dynamics and the state transition predictor, but the policy prior is trained in a supervised learning style and only used for regularizing the policy network. However, as shown in this paper, the policy can be exactly decoupled as these two parts without keeping an extra policy, where the state planner is optimized through policy gradient. \citet{edwards2020estimating} proposed a deterministic decoupled policy 
and updated the policy through a deterministic policy gradient similar to DDPG~\citep{lillicrap2015continuous}, based on the proposed $Q(s,s')$ rather than $Q(s, a)$; however, as revealed in their paper and our following test, the performance of $Q(s,s')$ is rather limited thus we omitted in our comparison. To sort out the difference between these methods and ours, we summarize the key factors in \tb{tb:inverse-dynamics-comparison}. In \ap{ap:related}, we discuss further works on transferable imitation learning.

%% file: papers/3-preliminary.tex
\section{Preliminaries}
\label{sec:pre}

\paragraph{Markov decision process.}

Consider a $\gamma$-discounted infinite horizon Markov decision process (MDP) $\caM = \langle \caS, \caA, \caT, \rho_0, r, \gamma \rangle$, where $\caS$ is the set of states, $\caA$ is the action space, $\caT: \caS \times \caA \times \caS \rightarrow [0, 1]$ is the environment dynamics distribution, $\rho_0: \caS\rightarrow[0,1]$ is the initial state distribution, and $\gamma\in [0,1]$ is the discount factor. The agent makes decisions through a policy $\pi(a|s): \caS \times \caA \rightarrow [0, 1]$ and receives rewards $r: \caS \times \caA \rightarrow \mathbb{R}$.
In our paper, we will assume the environment dynamics $\caT$ is a deterministic function such that $s' = \caT(s,a)$, and can have redundant actions, i.e., the transition probabilities can be written as linear combination of other actions'. Formally, this refers to the existence of a state $s_m\in\mathcal{S}$, an action $a_n\in\mathcal{A}$ and a distribution $p$ defined on $\mathcal{A}\setminus\{a_n\}$ such that $\int_{\mathcal{A}\setminus \{a_n\}}p(a)\mathcal{T}(s'|s_m, a)\dif a=\mathcal{T}(s'|s_m, a_n)$. We call two MDPs $\caM_1$ and $\caM_2$ share the same state transition when $\forall s,s' \in \caS_1=\caS_2=\caS$, $\exists a_1\in\caA_1, a_2\in\caA_2$ such that $s' = \caT_1(s,a_1) = \caT_2(s,a_2)$, and $\caA_1\neq\caA_2$ suggests different action dynamics. This constructs an important assumption for the transferring challenges in this paper.


\paragraph{Occupancy measure.}
The concept of occupancy measure (OM)~\citep{ho2016generative} is proposed to characterize the statistical properties of a certain policy interacting with a MDP. Specifically, the state OM is defined as the time-discounted cumulative stationary density over the states under a given policy $\pi$: $\rho_{\pi}(s) = \sum_{t=0}^{\infty}\gamma^t P(s_t=s|\pi)$.
Following such a definition we can define different OM:

a) State-action OM: $\rho_{\pi}(s,a) = \pi(a|s)\rho_{\pi}(s)$

\vspace{3pt}
\hbox{b) State transition OM: $\rho_{\pi}(s,s') = \int_{\caA}\rho_{\pi}(s,a)\caT(s'|s,a)\dif a$}

\vspace{-2pt}
c) Joint OM: $\rho_{\pi}(s,a,s') = \rho_{\pi}(s,a)\caT(s'|s,a)$


\paragraph{Imitation learning from state-only demonstrations.}

Imitation learning (IL) \citep{hussein2017imitation} studies the task of learning from demonstrations (LfD), which aims to learn a policy from expert demonstrations without getting access to the reward signals. The expert demonstrations typically consist of expert state-action pairs. General IL objective minimizes the state-action OM discrepancy:
\begin{equation}\label{eq:il}
\begin{aligned}
\pi^* &= \argmin_\pi \mathbb{E}_{s \sim \rho_{\pi}^{s}}\left [\ell \left (\piE(\cdot | s), \pi(\cdot | s)\right )\right ] \\
&\Rightarrow \argmin_\pi  \ell \left (\rho_{\piE}(s,a), \rho_{\pi}(s,a)\right )~,
\end{aligned}
\end{equation}
where $\ell$ denotes some distance metric. For example, GAIL~\citep{ho2016generative} chooses to minimize the JS divergence $\jsd(\rho_{\piE}(s,a) \| \rho_{\pi}(s,a))$, and AIRL~\citep{fu2018learning} utilizes the KL divergence $\kld(\rho_{\piE}(s,a) \| \rho_{\pi}(s,a))$, which also corresponds to a maximum entropy solution with the recovered reward~\citep{liu2020energy}. However, for the scenario studied in this paper, the action information is absent in demonstrations, which prevents usage of typical IL solutions. A popular solution \citep{torabi2019adversarial} is to instead optimize the discrepancy of the state transition OM with the state-to-action policy $\pi(a|s)$ as
\begin{equation}\label{eq:soil}
\pi^* = \argmin_\pi [\ell \left (\rho_{\piE}(s,s'), \rho_{\pi}(s,s')\right )]~.
\end{equation}

%% file: papers/4-method.tex
\section{Decoupled Policy Optimization Framework}
\begin{figure}[tbp]
     \centering
    \includegraphics[width=0.99\columnwidth]{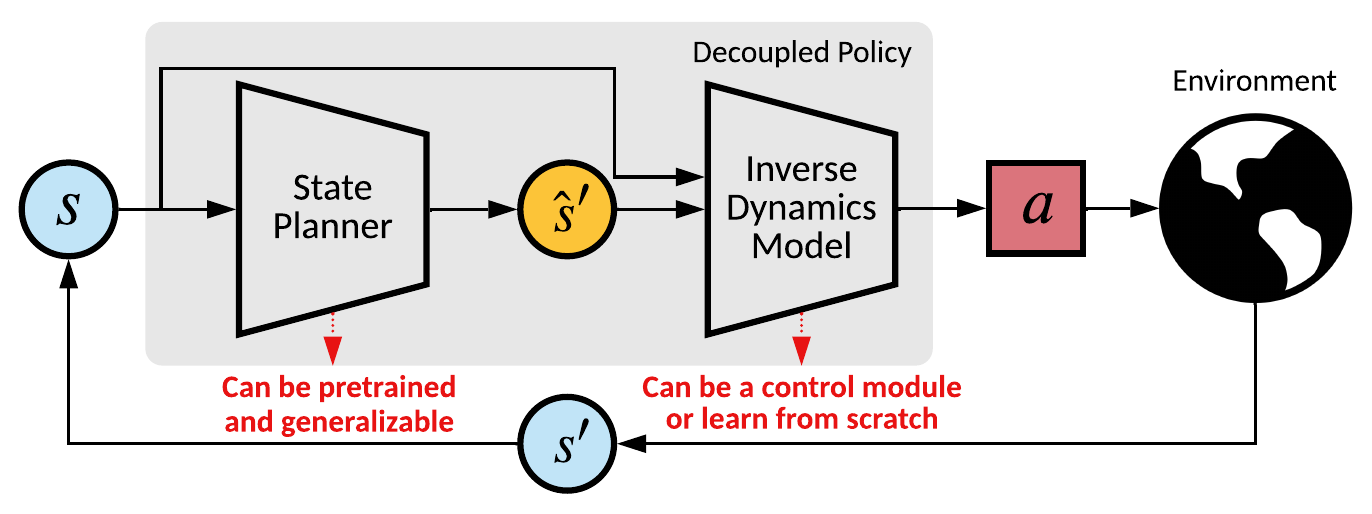}
    \vspace{-3pt}
    \caption{The architecture of Decoupled Policy Optimization (DePO), which consists of a state planner (to plan where to go) followed by an inverse dynamics model (to decide how to reach).}
    \vspace{-3pt}
    \label{fig:DePO}
\end{figure}

Previous works on SOIL focus on learning appropriate actions are ad-hoc for a specific task, ignoring understanding the high-level target planning strategy thus limiting the generalized ability. 
In this section, we present a novel architecture called Decoupled Policy Optimization (DePO) to recover a planner that predicts the neighboring targets where the expert aims to reach. The interaction with the environment further depends on a control module that can be formulated as an inverse dynamics model. An overview of the whole architecture is illustrated in \fig{fig:DePO}. To make the framework work as desired, we first formalize the decoupled policy; then, we introduce how to learn a generalized planner 
via both supervised learning and decoupled policy gradient; to alleviate the agnostic problem of the learned state plan, we further propose calibrated decoupled policy gradient to obtain a generalizable and accurate state planner.

\subsection{Decoupling the Policy}
In the formulation of SOIL, with absent information of actions, the optimality is defined via the state-transition OM. Unlike in standard IL tasks where the expert actions are accessible and perfectly imitating the expert policy corresponds to matching the state-action OM due to the one-to-one correspondence between $\pi$ and $\rho_{\pi}(s, a)$~\citep{ho2016generative,syed2008apprenticeship}, \prop{prop:one-to-more} states that finding a policy matching this objective is ambiguous and indirect.
\begin{proposition}\label{prop:one-to-more} 
  Suppose $\Pi$ is the policy space and $\mathcal{P}$ is a valid set of state transition OMs such that $\mathcal{P}=\{\rho: \rho\geq 0\newline\text{  and  }\exists\pi\in\Pi$, $\text{  s.t.  } \rho(s, s')=$ 
  $\rho_0(s)\int_{a}\pi(a|s)\caT(s'|s, a)\dif a\newline+\int_{\tilde{s}, a} \pi(a|s)\caT(s'|s, a)\rho(\tilde{s}, s)\dif \tilde{s}\dif a\}$, then a policy $\pi\in\Pi$ corresponds to one state transition OM $\rho_{\pi}\in\mathcal{P}$. Instead, under the action-redundant assumption about the dynamics $\caT$, a state transition OM $\rho\in\mathcal{P}$ can correspond to more than one policy in $\Pi$.
\end{proposition}
The proof is in \ap{ap:proofs}. 
In practical optimization, ambiguity is not really a critical issue since gradient descent-based algorithms have the ability to converge to one of multiple optima.
Nevertheless, without explicit modeling the expert's target planner (state transition), learning such an optimal state-to-action mapping policy is ad-hoc.
To enable the agent learn a high-level planning strategy from demonstrations that can be transferred to various low-level action spaces, we decouple the policy structure by a planner (formulated by a state transition planner) and a control module (formulated by an inverse dynamics model) by finding a one-to-one corresponding solution for SOIL. Before continuing, we introduce the definition of \textit{hyper-policy}.
\begin{definition}
  A hyper-policy $\Omega\in\Lambda$ is a maximal set of policies sharing the same state transition occupancy such that for any $\pi_1, \pi_2 \in\Omega$, we have $\rho_{\pi_1}(s,s')=\rho_{\pi_2}(s,s')$.
\end{definition}
Then by definition, \prop{prop:one-to-one} shows the one-to-one correspondence between $\Omega$ and $\rho_\Omega(s,s')$. Similar to the normal state-to-action mapping policy, a hyper-policy $\Omega$ can be represented as a state-to-state mapping function $h_{\Omega}(s'|s)$ which predicts the state transition such that for any $\pi\in\Omega$,
\begin{equation}\label{eq:state-transition}
\begin{small}
    h_{\Omega}(s'|s) =\frac{\rho_{\Omega}(s,s')}{\int_{\tilde{s}}\rho_{\Omega}(s,\tilde{s})\dif \tilde{s}}
    = \int_{a} \pi(a|s)\caT(s'|s,a)\dif a~.
\end{small}
\end{equation}
\vspace{-10pt}
\begin{proposition}\label{prop:one-to-one}
  Suppose the state planner $h_{\Omega}$ is defined as in \eq{eq:state-transition} and $\Gamma=\{h_{\Omega}: \Omega \in\Lambda\}$ is a valid set of the state transition predictor, $\mathcal{P}$ is a valid set of the state-transition OM defined as in \prop{prop:one-to-one}, then
  a state planner $h_{\Omega}\in\Gamma$ corresponds to one state transition OM, where $\pi\in\Omega$; and a state transition OM $\rho\in\mathcal{P}$ only corresponds to one hyper-policy state planner such that $h_{\rho} = \rho(s,s')/\int_{\tilde{s}}\rho(s,\tilde{s})\dif \tilde{s}$.
\end{proposition}
The proof can follow the Theorem 2 of \citet{syed2008apprenticeship}, and for completeness, we include it in \ap{ap:proofs}. Therefore, we find a one-to-one correspondence between the optimization term $\rho(s,s')$ and a state transition planner $h_{\Omega}(s'|s)$, which indicates that under state-only demonstrations we only need to recover the state transition prediction of the hyper-policy $\Omega_E$ as
\begin{equation}\label{eq:soil-hyper-policy}
\begin{aligned}
&\argmin_{\Omega} \ell \left (\rho_{\Omega_E}(s,s'), \rho_{\Omega}(s,s')\right ) \\
\Rightarrow &\argmin_{h_{\Omega}} \bbE_{s\sim\Omega}[\ell \left (h_{\Omega_E}(s'|s), h_{\Omega}(s'|s)\right )]
~.
\end{aligned}
\end{equation}
Nonetheless, we still require to learn a policy to interact with the MDP environment to match the state transition OM of the expert. This can be achieved by learning any policy $\pi\in\Omega_E$ according to \eq{eq:soil-hyper-policy} instead of recovering the expert policy $\piE$ exactly.

Intuitively, the state planner tells the agent the \textit{target} that the expert will reach without informing any feasible \textit{skill} that require the agent to learn.
Therefore, to recover a $\pi\in\Omega_E$, we can construct an inverse dynamics such that
\begin{equation}
    \pi = \underbrace{\caT^{-1}_{\pi}}_{\text{inverse dynamics}}(\underbrace{\caT(\piE)}_{\text{state planner}})~.
\end{equation}
Formally, the expert policy can be decoupled as
\begin{equation}\label{eq:decoupled-policy}
\begin{aligned}
    \piE(a|s) &= \int_{s'} \caT(s'|s,a)\piE(a|s)\dif s'\\
    &= \int_{s'} \frac{\rho_{\piE}(s,s')I_{\piE}(a|s,s')}{\rho_{\piE}(s)}\dif s'\\
    &= \int_{s'} h_{\piE}(s'|s) I_{\piE}(a|s,s')\dif s'~,
\end{aligned}
\end{equation}
where both the state planner $h$ and the inverse dynamics model $I$ are policy dependent. When the environment dynamics is deterministic, we have a simpler form on the probability of taking the action $a$ as
\begin{equation}\label{eq:det-decoupled-policy}
\begin{aligned}
\piE(a|s) &= h_{\piE}(s'|s) I_{\piE}(a|s,s').
\end{aligned}
\end{equation}
In optimality, we should have $s'=\caT(s,a)$, yet in the inference stage we actually first predict $\hat{s}'$ from $h$ and then take the action $a$.
Fortunately, the optimality in SOIL only requires us to recover $\pi\in\Omega_E$, we do not have to learn about $I_{\piE}$ but just one feasible skill $I(a|s,s')$. Then the policy distribution given the state $s$ can be recovered by
\begin{equation}\label{eq:hyper-pie-decouple}
\begin{aligned}
    \pi(\cdot|s) = \bbE_{\underbrace{{\scriptsize \hat{s}'\sim h_{\Omega_E}(\hat{s}'|s)}}_{\normalsize \text{\scriptsize target}}} \Big[ \underbrace{I(\cdot|s,\hat{s}')}_{\text{skill}} \Big]~.
\end{aligned}
\end{equation}
Here the inverse dynamics model $I$ serves as a control module that offers an arbitrary \textit{skill} to reach the expected \textit{target} state provided by the state planner $h$, and does not depend on the hyper-policy $\Omega_E$. In the cases where it must be learned from scratch, we need a sampling policy $\pi_{\caB}$ to construct $I=I_{\pi_{\caB}}$, and a mild requirement for $\pi_{\caB}$ that it covers the support of $\rho_{\Omega_E}(s,s')$ so that the learned $I$ can provide a possible action to achieve the target state. 
Furthermore, if the environment and the expert policy are both deterministic, the state transition is a Dirac delta function, and $h$ is also a simple deterministic function.

\subsection{Supervised Learning from Data}
We provide a quick view on the simplest way to learn both modules by directly supervised learning from data.
\paragraph{Inverse dynamics model.}
The requirement of interacting with the environment asks for a control module for reaching the target predicted by the planner, which can be formulated as an inverse dynamics model predicting the action given two consecutive states. Formally, let the $\phi$-parameterized inverse dynamics model $I_{\phi}$ take the input of a state pair and predict a feasible action to achieve the state transition: $\hat{a} = I_{\phi}(s,s')$.
We note that the inverse dynamics model can either be pre-trained in advance, or a prior ground-truth function, or any differentiable control module that plays the role of planning. If we must learn the model online, we can choose to minimize the divergence (for example, KL) between the inverse dynamics of a sampling policy $\pi_{\caB}$ and $I_{\phi}$, i.e.,
\begin{equation}\label{eq:inverse-dynamics}
    \min_{\psi}L^{I} = \bbE_{(s,s')\sim\pi_{\caB}}[\fd( I_{\pi_{\caB}}(a|s,s')\| I_{\phi}(a|s,s'))]~.
\end{equation}
The accuracy of this module is important, otherwise, the agent cannot take specific action to reach the target state. Fortunately, recall that we only need the support of learned $I(a|s,s')$ to cover the support of the expert state transition OM at convergence, from which we can infer at least one possible action to accomplish the transition. 
Besides, due to the principle of the inverse dynamics serving as a local control module, when learning from experience data, the model only has to focus on inferring accurate actions on states encountered by the current policy instead of the overall state space. Thence, in practice, we train the inverse dynamics model every time before updating the policy. 

\paragraph{State planner.}
The state planner is constructed as modeling the explicit information of the expert by predicting the subsequent state of the expert under the current state. Thus, to learn a parameterized module $h_{\psi}$, a direct way is to match the expert data as minimizing the divergence to achieve \eq{eq:soil-hyper-policy}:
\begin{equation}\label{eq:sup-sp}
    \min_{\psi}L^{h} = \bbE_{(s,s')\sim\Omega_E}[\fd( h_{\Omega_E}(s'|s)\| h_{\psi}(s'|s))]~.
\end{equation}
However, simply mimicking the state-transition in dataset tends to result in serious compounding error problems, as behavior cloning (BC) in standard imitation learning~\cite{ross2011reduction}. 
To alleviate this problem, we can resort to the help of policy gradient and incorporate generative adversarial objective into the decoupled policy learning.

\subsection{Decoupled Policy Gradient}
\label{sec:depg}
By now, we have explained why and how to decouple the policy as two modules, and they are naturally coherent as an integral policy function, thereafter, with a well-defined inverse dynamics, we can derive policy gradient (PG) for optimizing the high-level state planner, named Decoupled PG (DePG) . 
The predicted target states by the planner will further drive the inverse dynamics to provide desirable actions leading to match the expert state-transition OM. 

Assume we can obtain the reward of the task, and denote the state-action value function as $Q(s,a)$, then by chain rule, DePG can be accomplished by
\begin{align}
    \nabla_{\phi, \psi} \caL^\pi &= \bbE_{(s,a)\sim\pi}\left [Q(s,a)\nabla_{\phi,\psi}\log{\pi_{\phi,\psi}(a|s)}\right ] \nonumber \\
    &= \bbE_{(s,a)\sim\pi}\Big[\frac{Q(s,a)}{\pi(a|s)}\Big(\int_{s'}I(a|s,s')\nabla_{\psi}h_{\psi}(s'|s)\dif s' \nonumber \\
    &\quad+\bbE_{s'\sim h}\left [\nabla_{\phi}I_{\phi}(a|s,s')\right ]\Big)\Big]~. \label{eq:pg}
\end{align}

In our formulation, since we have assumed the inverse dynamics to be an accurate control module at least for the current policy, the inverse dynamics function $I$ is static when optimizing the policy function. As such, the gradient form in \eq{eq:pg} only updates $\psi$ (the state planner) and the gradient on $\phi$ (the inverse dynamics) should be dropped. In addition, since the state planner is approximated by neural networks (NNs) in our case, it is convenient to apply the reparameterization trick and bypass explicitly computing the integral over $s'$ as
\begin{equation}
\begin{aligned}\label{eq:repara}
    s' = h(\epsilon;s), ~~
    \pi(a|s)&=\bbE_{\epsilon\sim \mathcal{N}}\left [ I(a|s,h(\epsilon;s)) \right ]~,
\end{aligned}
\end{equation}
where $\epsilon$ is an input noise vector sampled from some fixed distribution, like a Gaussian.
Then \eq{eq:pg} becomes
\begin{small}
\begin{equation}
\begin{aligned}\label{eq:pg-sp}
    \nabla_{\psi} \caL^\pi
        &= \bbE_{(s,a)\sim\pi,\epsilon\sim\mathcal{N}}\left [\frac{Q(s,a)}{\pi(a|s)} \big(\nabla_{h}I(a|s,h_\psi(\epsilon;s))\nabla_{\psi}h_{\psi}(\epsilon;s)\big)\right ]~.
\end{aligned}
\end{equation}
\end{small}
\noindent In this form, we are taking the knowledge from the inverse dynamics about action $a$ to update parameters of the state planner by updating the prediction about the next state with error $\Delta s' = \alpha \nabla_{h}I(a|s,h(\epsilon;s))$ and learning rate $\alpha$. 

\paragraph{Agnostic decoupled PG.} 
Till now, we take for granted the existence of an accurate control module for the current policy. This makes sense if we have the ground-truth inverse dynamics function of the environment. In more general scenarios, the skill model has to be learned during the training stage. 
Unfortunately, simply applying DePG (\eq{eq:pg}) to obtain the desired high-level target planner is faced with serious learning challenges, especially when the inverse dynamics model is approximated by NNs.

The problem comes from a constraint on the inverse dynamics modelling's inputs, which requires the prediction of the target planner to be a legal neighbor state that follows the inputs training data distribution.
Otherwise the state transition pair $(s, s')$ will be an illegal transition 
and leads the corresponding output actions to be agnostic.
An illegal state transition could still be a legal input to the approximated inverse dynamics model if it may still provide a feasible action to interact with the environment.
However, since we are utilizing the gradient of the inverse dynamics model $\nabla_{h}I(a|s,h_\psi(\epsilon;s))$,
we do not expect a generalization on illegal transitions to get a feasible action. 
In other words, simply optimizing $\psi$ through \eq{eq:pg}, we cannot constrain the gradient provided by the inverse dynamics within pointing to a reachable target state, and the state planner will be trained to predict an arbitrary and unreasonable state that can still lead the inverse dynamics model to give a feasible action. This is further revealed in \se{sec:synthetic}.

\paragraph{Calibrated decoupled PG.} To alleviate such a problem, we should dive into the nature of the decoupled structure. The goal of DePO is to learn a planner that determines the high-level target to reach, yet taking which action to achieve is inessential. In the following theorem, we draw insights on regularizing the planner to provide a feasible state plan.  
\setcounter{theorem}{0}
\begin{theorem}
    Let \caA(s) be an action set that for all $a$ in $\caA(s)$, a deterministic transition function leads to the same state $s'=\caT(s,a)$. If there exists an optimal policy $\pi^*$ and a state $\hat{s}$ such that $\pi^*(\cdot|\hat{s})$ is a distribution over $\caA(\hat{s})$, then we can replace $\pi^*(\cdot|\hat{s})$ with any distributions over $\caA(\hat{s})$ which does not affect the optimality.
\end{theorem}

The proof is in \ap{ap:proofs}.
Therefore, no matter what action $a$ the inverse dynamics takes, as long as it leads the agent from $s$ to $s'$, the integrated policies are all optimal policies on $s$ and their value $Q_{\pi}(s,a)$ should be the same. On this condition, give a state-action pair $(s,a)$ and a legal transition $s'$, we can simplify the inverse dynamics as a deterministic distribution on $a$ with the probability of one, i.e., $I(a|s,s')=1$, then the probability of the decoupled policy $\pi(a|s)=h(s'|s)$, where $s'$ is the target state, and DePG (\eq{eq:pg}) can be simplified as
\begin{equation}\label{eq:weightedmle-sp}
    \begin{aligned}
        \nabla_{\psi} \caL^\pi
        &= \bbE_{(s,a,s')\sim\pi}\left [Q(s,a) \nabla_{\psi}\log h_{\psi}(s'|s)\right ]~,
    \end{aligned}
\end{equation}
which is named Calibrated Decoupled PG (CDePG). Optimizing with CDePG can be realized as maximizing the probability to target state $s'$ on state $s$ if $a$ is a good action regarding the inverse dynamics is accurate. However, solely updating \eq{eq:weightedmle-sp} leads to a severe exploration problem since the planner is only allowed to predict a visited state. Fortunately, DePG provides a way to explore the most promising actions although it is not responsible for getting legal state transition. Thereafter, in practice, we choose to jointly optimize DePG (\eq{eq:pg-sp}) with CDePG (\eq{eq:weightedmle-sp}) and find it can achieve a good balance between exploration and accurate prediction on legal targets. 
More detailed analyses are provided in \ap{ap:ablation}. 
Note that our derivation does not only limit in the literature of SOIL but also can be applied to general RL tasks. Yet in this paper, we focus on SOIL tasks since our motivation is to understand the high-level planning strategy of experts.

\paragraph{Generative adversarial training.}
To reach the optimality, the inverse dynamics model must converge to cover the support of the expert hyper-policy, which is essential especially when learning the inverse dynamics from scratch. This asks for the data-collecting policy to sample around the expert's occupancy. 
This is easy to achieve on simple low-dimensional tasks, but may not be satisfied in high-dimensional continuous environments. Also, due to the limited amount of the demonstration data, simply supervised learning the state planner module can cause serious compounding error due to the one-step optimization and the agent may not know where to go in an unseen state.
To this end, we incorporate the GAN-like informative rewards similar to GAIfO~\citep{torabi2019adversarial} and update the policy using the induced decoupled PGs.
In detail, we construct a parameterized discriminator $D_\omega(s,s')$ to compute the reward $r(s,a)\triangleq r(s,s')$ as $\log{D_\omega(s,s')}$ while the decoupled policy serves as the generator.

The optimization can be conducted using any PG-based RL learning algorithms (e.g., TRPO, PPO, SAC).
As the training proceeds, we expect the agent to sample more transition data around $\Omega_E$, and thus the support of the sampling policy progressively covers the support of $\rho_{\Omega_E}(s,s')$ and the agent can generalize well on unseen states. 

\subsection{Overall Algorithm} 

The algorithm is composed with three essential parts: the state planner $h$ used for predicting next states; the inverse dynamics model $I$ used for inferring the possible actions conditioned on two adjacent states; and the discriminator $D$ offers intermediate reward signals for training the decoupled policy $\pi$. When the inverse dynamics model $I$ is obtained in advance, we only have to learn $h$; yet when we need to learn both modules from scratch: first learn $I$ and then train $h$ with $I$ fixed.
The overall objective of DePO is
\begin{align}
    \min_{\phi} &~ L^{I}~~\text{(if $I$ needs to be learned)},\label{eq:final-loss-inv}\\
    \min_{\psi} &~ \caL^{\pi,h} = \caL^{\pi} + \lambda_h \caL^h ~,
    \label{eq:final-loss}
\end{align}
where $\lambda_h$ is the hyperparameter for trading off the loss. The detailed algorithm is summarized in \ap{ap:algo}. The flexibility of DePO comes from the decoupled structure, such that the state planner can be transferred to different MDPs  without training if they share the same state transition, and only the inverse dynamics module needs to be retrained, which yields significant gain of sample efficiency as shown in the experiments (\se{sec:exps}).

\subsection{Analysis of The Compounding Error}
\label{sec:compounding-error}

In our formulation, we have decoupled the state-to-action mapping policy as a state-to-state mapping function and a state-pair-to-action mapping function. This introduces a new compounding error challenge such that the agent cannot reach where it plans due to the generalization errors of these two modules.

\begin{theorem}[Error Bound of DePO]\label{theorem:compounding-error}
    Consider a deterministic environment whose dynamics transition function $\caT(s,a)$ is deterministic and $L$-Lipschitz. Assume the ground-truth state transition $h_{\Omega_E}(s)$ is deterministic, and for each policy $\pi\in\Pi$, its inverse dynamics $I_{\pi}$ is also deterministic and $C$-Lipschitz. Then for any state $s$, the distance between the desired state $s'_E$ and reaching state $s'$ sampled by the decoupled policy is bounded by
\begin{equation*}
    \|s'-s'_E\| \leq LC \| h_{\Omega_E}(s) - h_{\psi}(s) \| + L\|I_{\pi_{\caB}}(s,\hat{s}') - I_{\phi}(s,\hat{s}')\|~,
\end{equation*}
where $\pi_{\caB}$ is a sampling policy that covers the state transition support of the expert hyper-policy and $\hat{s}' = h_{\psi}(s)$ is the predicted next state.
\end{theorem}

The proof can be found in \ap{ap:proofs}. From \theo{theorem:compounding-error} we know that the compounding error can be enlarged due to each part's generalization error, where the first term corresponds to the error of predicted states and the second term indicates whether the agent can reach where it plans to.

%% file: papers/5-exp.tex
\section{Experiments}
\label{sec:exps}


In this section, we conduct in-depth analyses of our proposed DePO method. We first conduct a simple 2D grid world environment and validate the planning ability of DePO on out-of-demonstration state regions (\se{sec:synthetic}); then, we show that DePO can achieve the best imitation performance compared with baselines (\se{sec:evaluation}). Afterwards, we investigate intriguing usages of DePO, which includes transferring by pre-training (\se{sec:transfer}), and co-training (\se{sec:cotrain}) for agents over various action spaces or dynamics.
Due to the page limit, we leave more details, additional results and ablation studies in \ap{ap:exps}.

\subsection{Generalized Planning Ability}
\label{sec:synthetic}

\begin{figure*}[!t]
\centering
\subfigure[Grid World Setting]{
\begin{minipage}[b]{0.21\linewidth}
\centering
\label{fig:expert_heat}
\includegraphics[width=0.85\linewidth]{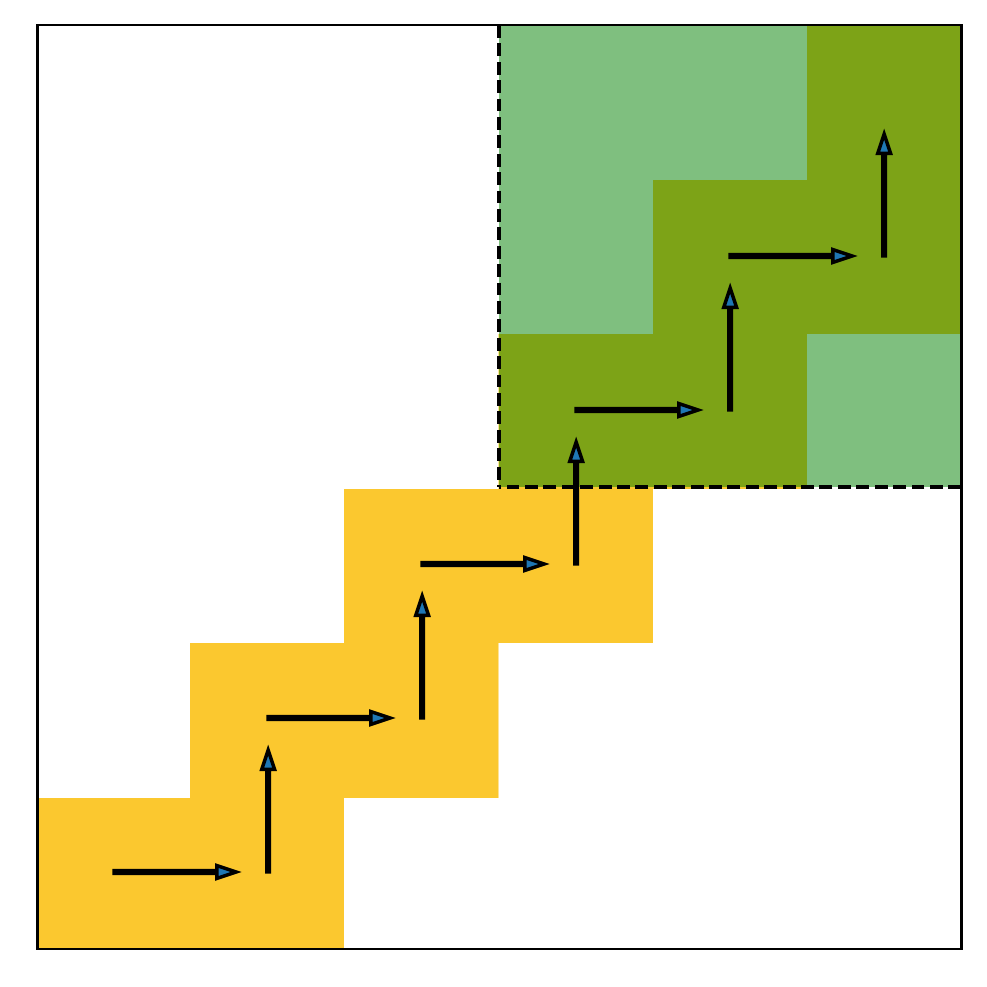}
\end{minipage}
}
\subfigure[Supervised DePO]{
\begin{minipage}[b]{0.21\linewidth}
\centering
\label{fig:ebil_heat}
\includegraphics[width=0.85\linewidth]{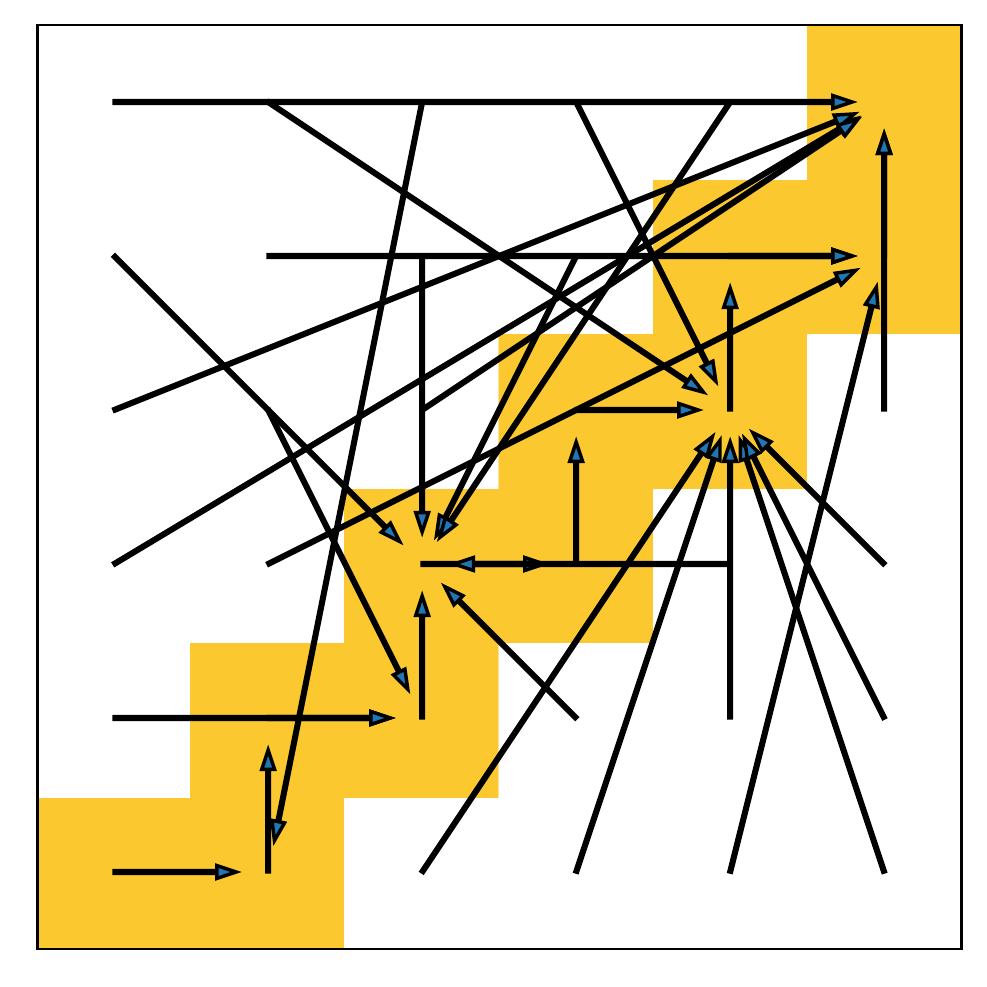}
\end{minipage}
}
\subfigure[Agnostic DePG]{
\begin{minipage}[b]{0.21\linewidth}
\centering
\label{fig:gail_heat}
\includegraphics[width=0.85\linewidth]{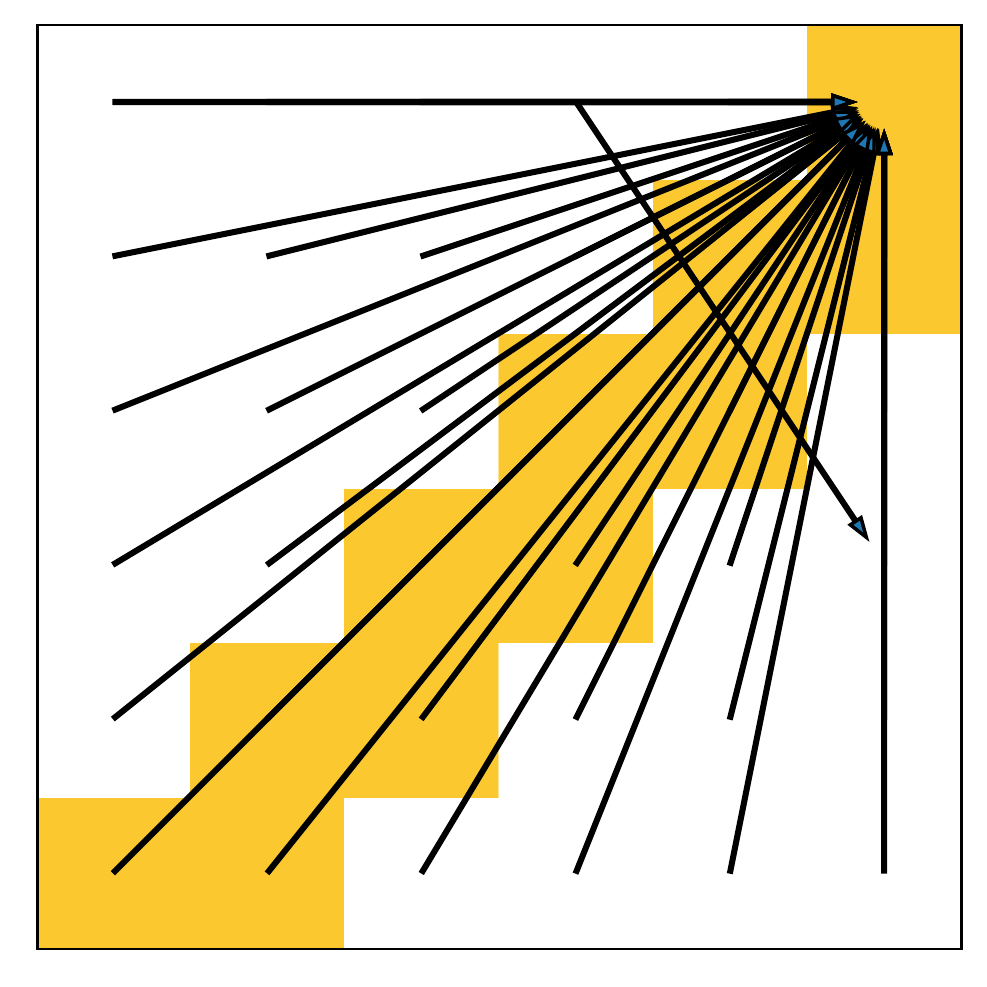}
\end{minipage}
}
\subfigure[DePO]{
\begin{minipage}[b]{0.21\linewidth}
\centering
\label{fig:airl_heat}
\includegraphics[width=0.85\linewidth]{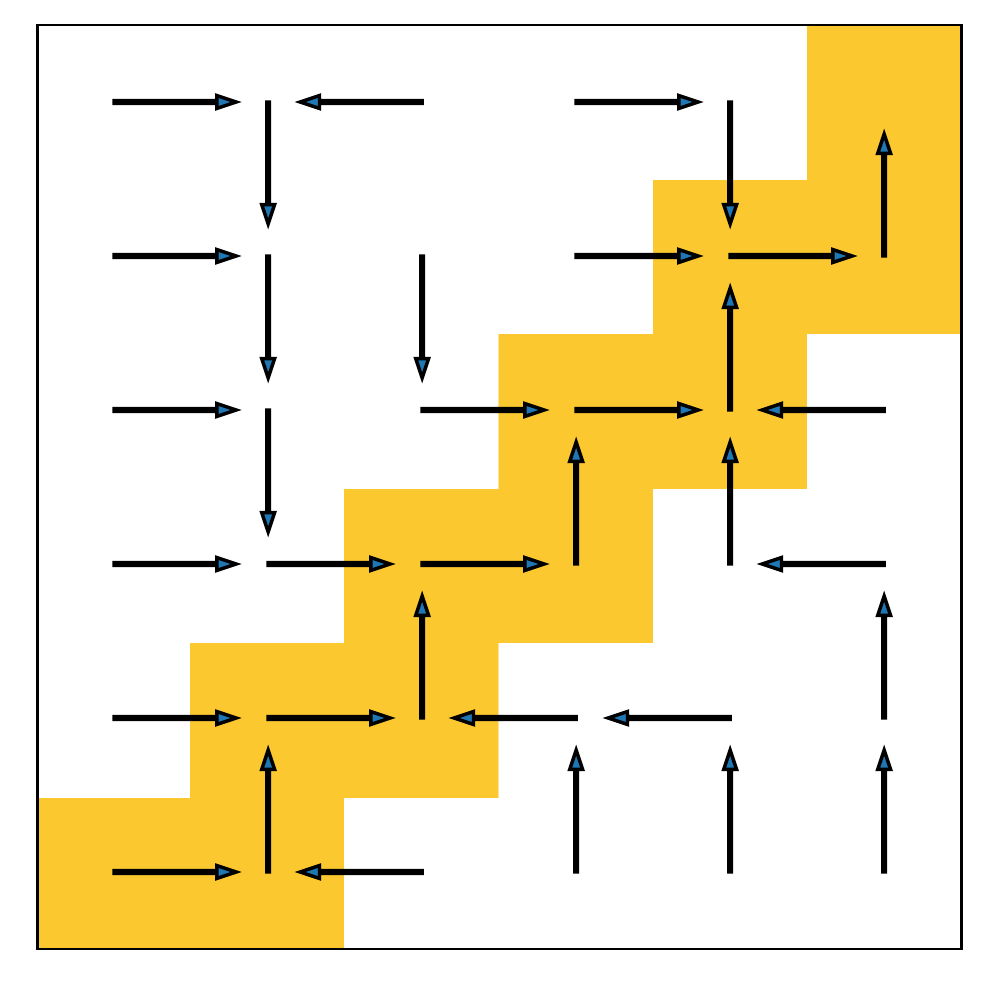}
\end{minipage}
}
\vspace{-10pt}
   \caption{Grid world environment and the prediction of the learned state planner. (a) The expert starts from the left-bottom corner (0,0) to the right-upper (6,6) and the arrows on the yellow grid depicts the path of the expert. The agent is required to start at any grid on the map except the shaded zone. We test three variants under the decoupled policy structure where both modules are learned from scratch. (b) Supervised learning only from the dataset results in predictions of the target state on expert paths, even if not a neighboring (legal) one. (c) Agnostic DePG learns to predict arbitrary states, 
   while the inverse dynamics can still give a legal action to reach a neighbor state. (d) The proposed DePO algorithm, which generalizes the planning into every out-of-demonstration state (white blocks) with legal transitions.}
\vspace{-10pt}
\label{fig:synthetic}
\end{figure*}

The key technical contribution of DePO includes the decoupled structure of policy that provides the ability to plan the state transition to match the expert. 
Therefore, we are keen to verify the non-trivial generalized planning ability, especially on out-of-demonstration states. We generate expert demonstrations in a 2D 6$\times$6 grid world environment, in which the expert starts at the left bottom corner (0,0) and aims to reach the upper right corner (5,5) (\fig{fig:synthetic}(a)). The agent starts at any grid randomly except the shaded zone, and in each grid, it has $k\times 4$ actions, which means the agent has $k$ available actions to reach a neighboring block. All functions are learned from scratch. We first employ simple supervised learning on the two modules, shown in \fig{fig:synthetic}(b), which indicates that the agent can only predict the target state on the demonstrated path, even if it is not a neighboring one; then, we illustrate the agnostic problem of DePG in \fig{fig:synthetic}(c), where it learns to predict arbitrary states, even if it is not a legal transition. However, with the CDePG proposed in \eq{eq:weightedmle-sp}, DePO shows great planning ability on the legal transition to match the expert occupancy measure, even on out-of-demonstration states.

\subsection{Imitation Evaluations}
\label{sec:evaluation}

\begin{figure*}[t]
    \centering
    \includegraphics[width=0.9\textwidth,height=3.3cm]{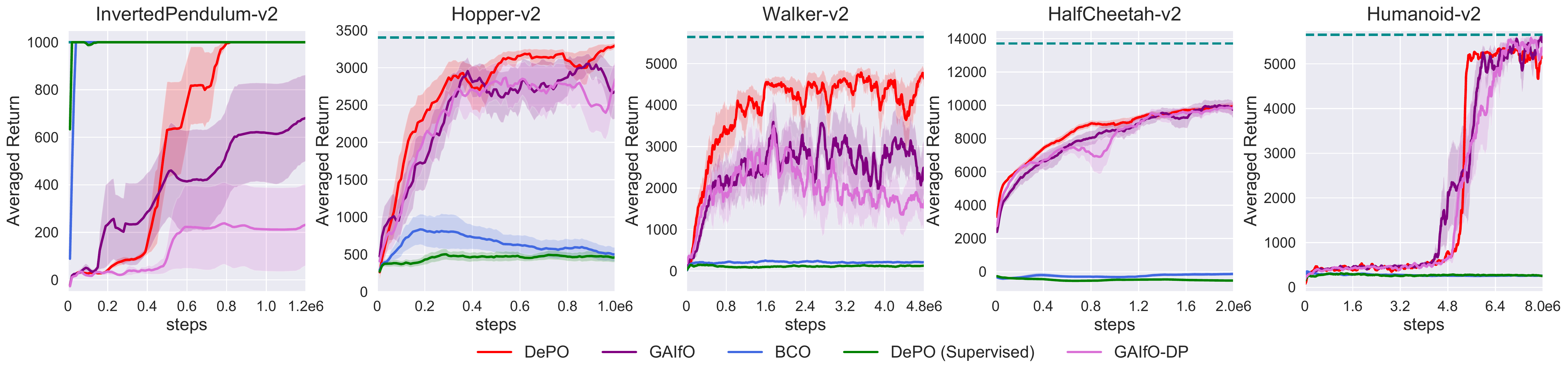}
    \vspace{-10pt}
    \caption{Learning curves on easy-to-hard continuous control benchmarks, where the dash lines represent the expert performance.}
    \vspace{-5pt}
    \label{fig:mujoco-curves}
\end{figure*}

We show that DePO keeps the best imitation performance by comparing against other baseline methods on easy-to-hard continuous control benchmarking environments (\fig{fig:mujoco-curves}).
In each environment, besides GAIfO and BCO, we also evaluate GAIfO with decoupled policy (denoted as GAIfO-DP) and the supervised learning version of DePO. For fairness, we re-implement all algorithms and adopt Soft Actor-Critic (SAC)~\citep{haarnoja2018soft} as the underlying RL learning algorithm for GAIfO and DePO.
For each environment, we train an SAC agent to collect 4 state-only trajectories as expert data. All algorithms are trained with the same frequency and gradient steps.

It is easily concluded that for simple environments as InvertedPendulum, supervised learning algorithms can achieve better learning efficiency since the state space is limited. However, on difficult tasks, BCO and DePO (Supervised) both fail due to compounding errors, yet policy gradient methods like DePO and GAIfO own better imitation efficacy, and DePO can gain the best performance against its counterparts. 
In \ap{ap:exps-res}, we reveal that DePO enables the agent to reach exactly where it plans to, and the state planner is even such accurate that it can be used for long (more than one)-step planning.
The corresponding curves of the errors between the predicted state and the reached state are further shown in \ap{ap:reach-prediction}. To demonstrate the planning ability of DePO, we illustrate the images of the imagined rollout states provided by the learned state planner, compared with the reaching states achieved during interaction with the environment, shown in \ap{ap:planner}.

\subsection{Few-Shot Transferring by Pre-training}
\label{sec:transfer}

\begin{figure}[tbp]
\centering
\includegraphics[width=0.78\linewidth]{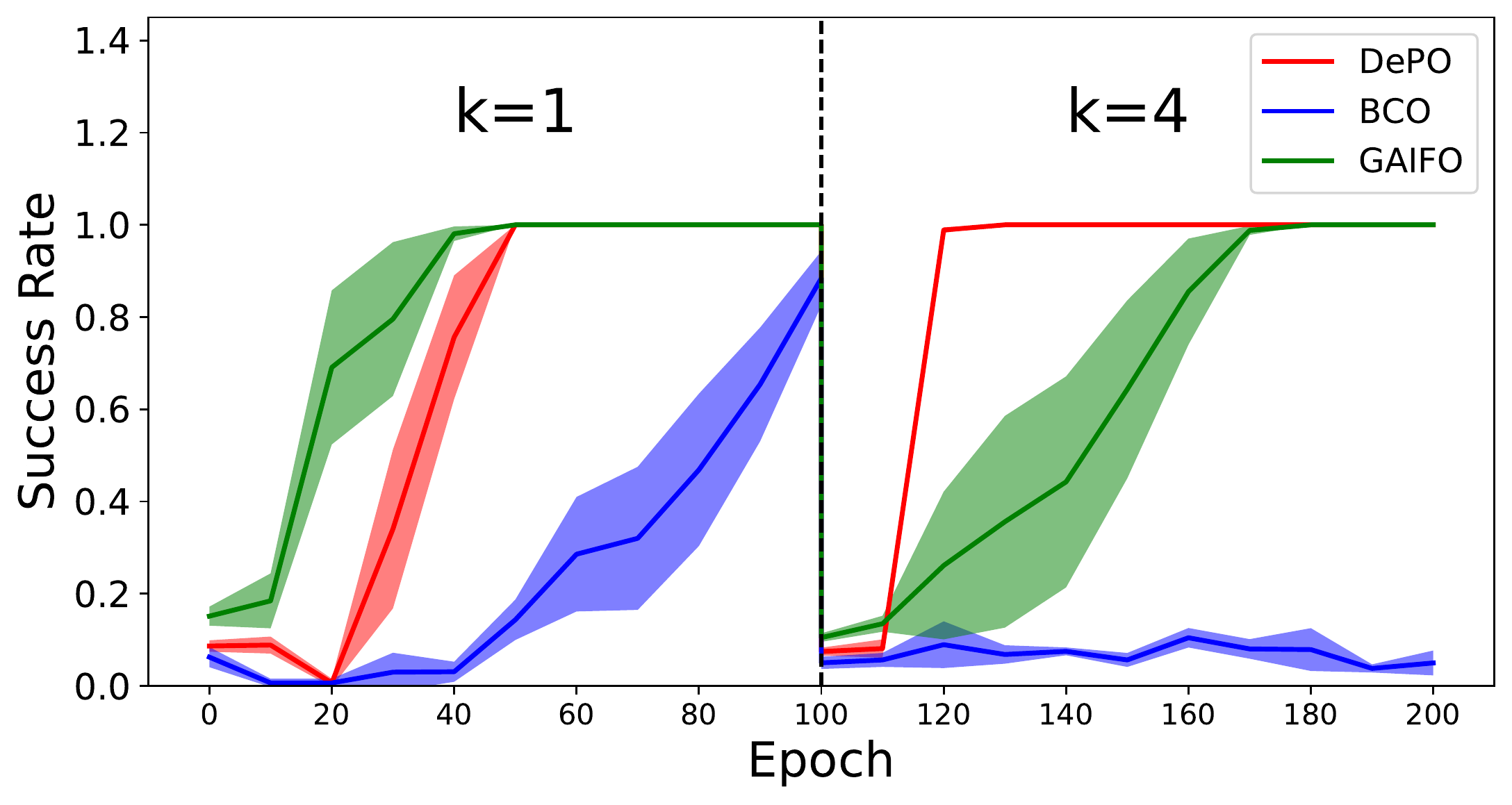}
\vspace{-10pt}
\caption{Transferring experiment on grid world environment. The y-axis denotes the success rate of reaching (5,5). The solid line and the shade shown in this and following figures represent the mean and the standard deviation of the results over 5 random seeds.}
\vspace{-12pt}
\label{fig:grid-transfer}
\end{figure}

\begin{figure*}[t]
    \centering
    \includegraphics[width=0.9\textwidth,height=2.95cm]{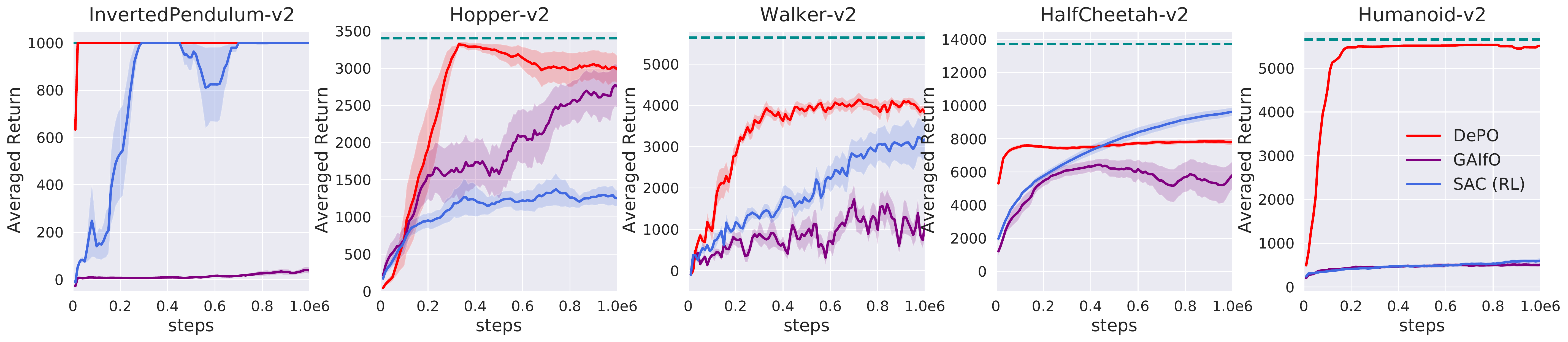}
    \vspace{-10pt}
    \caption{Transferring by pre-training on Mujoco tasks with complex action dynamics within 1e6 interaction steps.}
    \vspace{-10pt}
    \label{fig:mujoco-transfer}
\end{figure*}

We are curious about the potential usage of DePO in transferring the policy to a different skill space (e.g., different action space or different dynamcis) with a pre-trained state planner. We conduct the experiments on both the discrete grid-world task (\fig{fig:grid-transfer}) and the continuous control benchmarks (\fig{fig:mujoco-transfer}). 
In particular, on grid-world, we train each agent with the action space parameter $k=1$ for 100 epochs, then we reset the environment and change $k=4$ for the transferring training stage. On Mujoco tasks, we use the pre-trained state planner in previous imitation tasks (\se{sec:evaluation}) and start the training on a complex action dynamics (see \ap{ap:transfer-setting} for details) for the transferring stage.

For BCO and GAIfO, since the policy is a state-to-action mapping function, it must be re-trained from scratch and need much more exploration data. However, with the decoupled policy structure, DePO can keep the state-predictor and is only required to re-train the inverse dynamics model. Learning curves on all experiments reveal the fast sample efficiency and favorable stability in the transferring stage, which is even more efficient than online SAC RL agents (\fig{fig:mujoco-transfer}). Moreover, consider if we have a ground truth inverse dynamics function, DePO even does not need any training and can be directly deployed for tasks with different skills (see \ap{ap:transfer:ppuu}). This motivates an attractive application of DePO for pre-training in a simple task and deployed into a more complex skill space. And in \ap{ap:transfer:rl} we show DePO can be pre-trained on RL agents (without demonstration but reward instead) and is still able to be transferred to learn different skills with much higher efficiency compared with the state-of-the-art RL algorithm.

\subsection{Co-training and Transferring for Real-World Application}
\label{sec:cotrain}

The rapid development of autonomous driving has brought huge demand on high-fidelity traffic participants interactions simulation for close-loop training and testing \citep{zhou2020smarts}. 
However, driver's detailed actions are not easy to obtain, yet we adopt SOIL from a traffic surveillance recording dataset (NGSIM I-80~\citep{halkias2006next}) that contains kinds of recorded human driving trajectories. For less training time, we only utilize half of the dataset.
In addition, for simulating different kinds of vehicles (such as cars and trucks), we should train different policies due to the difference in the underlying controlling strategies of different vehicles. However, note that they share the high-level planning mode of feasible states transitions. To this end, we show another appealing usage of DePO by co-training agents with different action dynamics but learning one single shared state planner. 

\begin{figure}[htbp]
    \centering
    \includegraphics[width=0.99\columnwidth]{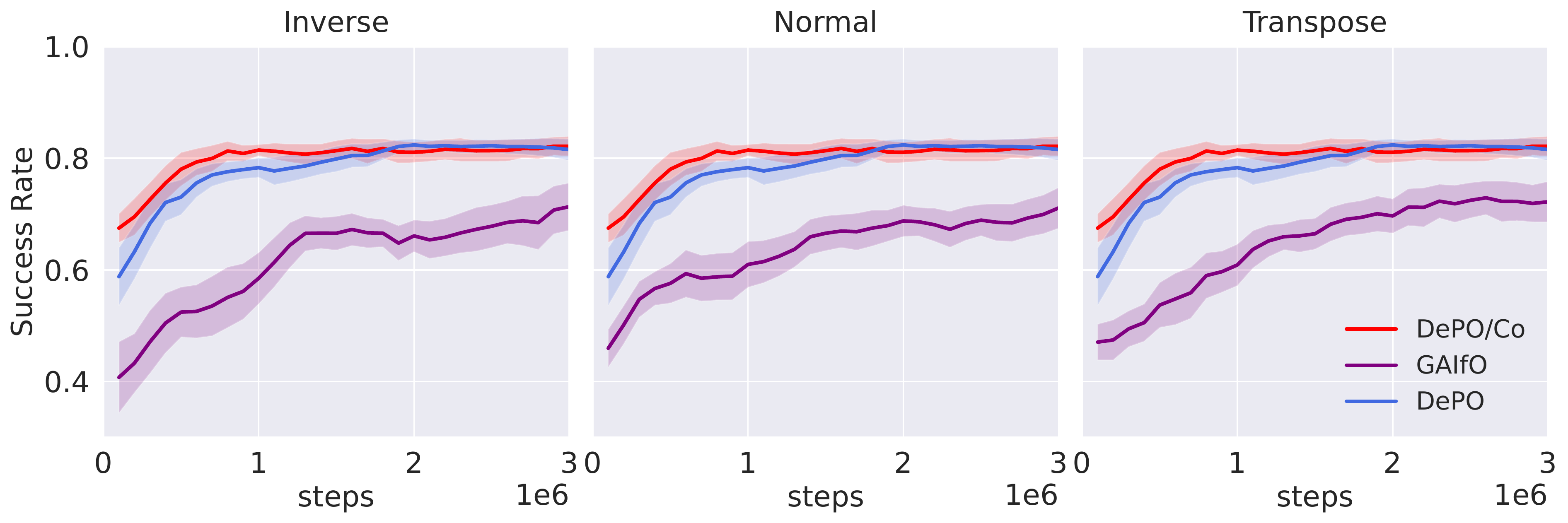}
    \vspace{-8pt}
    \caption{Co-training for different vehicles.}
    \vspace{-15pt}
    \label{fig:ppuu-cotrain}
\end{figure}

We utilize the simulator provided by \citet{henaff2019model} as our simulation platform.
Specifically, based on the original dynamics (which we refer to as \textit{Normal}), we design two additional action dynamics: \textit{Inverse} -- the action is taken additive inverse before stepping in the environment; and \textit{Transpose} -- the two dimensions of the action are swapped before stepping. In this experiment, we construct a direct inverse dynamics function without training, and compare DePO and GAIfO, which independently trains three different policies for this task, with the co-training version of DePO (denoted as DePO/Co). We plot the \textit{Success Rate} during the training in \fig{fig:ppuu-cotrain}, denoting the percentage of driving across the entire area without crashing into other vehicles or driving off the road among all evaluation vehicles. 
We observe DePO benefits a lot efficiency from learning the high-level planner with ground truth inverse dynamics functions; and DePO/Co can take advantage of the shared training of all agents, achieving much less sampling cost and the fastest convergence rate. Additional statistic results for more metrics and transferring experiments can be further referred to \ap{ap:ppuu-exps}.

%% file: papers/6-conclusion.tex
\section{Conclusion}
\label{sec:conclusion}

We propose Decoupled Policy Optimization (DePO) for transferable state-only imitation learning, which decouples the state-to-action mapping policy into a state-to-state mapping state planner and a state-pair-to-action mapping inverse dynamics model. 
DePO allows for adapting to different agents and generalizing on out-of-demo states with decoupled policy gradient and generative adversarial objective.
Extensive experiments demonstrate the power of DePO with appealing usages as transferring by pre-training and effective co-training for different skilled agents. For future works, we plan to generalize DePO to image-based problems, goal-conditioned tasks and apply to real-world robotics.

\section*{Acknoledgement}
We thank Cheng Chen, Jian Shen, Zhengyu Yang, Menghui Zhu, Hanye Zhao, Minkai Xu and Yuxuan Song for helpful discussions. The SJTU team is supported by ``New Generation of AI 2030'' Major Project (2018AAA0100900), Shanghai Municipal Science and Technology Major Project (2021SHZDZX0102) and National Natural Science Foundation of China (62076161). The work is also sponsored by Huawei Innovation Research Program.
The author Minghuan Liu is also supported by Wu Wen Jun Honorary Doctoral Scholarship, AI Institute, Shanghai Jiao Tong University. 

%% file: papers/7-appendix.tex
\newpage
\appendix

\section{Algorithm}
\label{ap:algo}

\begin{algorithm}[h]
    \caption{Decoupled Policy Optimization (DePO)}
    \label{alg:decoupled-policy}
    \begin{algorithmic}
       \STATE {\bfseries Input:} State-only expert demonstration data $\caD=\{ (s_i) \}_{i=1}^{N}$, empty replay buffer $\caB$, randomly initialized discriminator model $D_{\omega}$, state transition predictor $h_{\psi}$ and parameterized inverse dynamics model $I_{\phi}$
       
       \FOR{$k = 0, 1, 2, \cdots$} 
       \STATE{$\triangleright$ Pre-training stage}
        \STATE Collect trajectories $\{(s,a,s',\text{done})\}$ using a random initialized policy $\pi=\bbE_{\epsilon\sim \mathcal{N}}\left [ I_{\phi}(a|s,h_\psi(\epsilon;s)) \right ]$ and store in $\caB$
       \STATE Sample $(s,a,s')\sim\caB$ and update $\phi$ by $L^{I}$ (\eq{eq:inverse-dynamics})
       \ENDFOR
       
       \FOR{$k = 0, 1, 2, \cdots$}
       \STATE{$\triangleright$ Online training stage}
       \STATE Collect trajectories $\{(s,a,s',r,\text{done})\}$ using current policy $\pi=\bbE_{\epsilon\sim \mathcal{N}}\left [ I_{\phi}(a|s,h_\psi(\epsilon;s)) \right ]$, where $r$ can be obtained from the environment (RL) or the discriminator $D_\omega$ (IL), and store in $\caB$
       \STATE Sample $(s,a,s')\sim\caB, (s,s')\sim\caD$
        \IF{\textit{learn inverse dynamics function}}
        \REPEAT
       \STATE Update $\phi$ by $L^{I}$ (\eq{eq:inverse-dynamics})
       \UNTIL{Converged}
       \ENDIF
       \IF{\textit{doing imitation learning}}
       \STATE Update the discriminator $D_\omega$ with the loss:
       $$
            \caL_{\omega}^D = -\bbE_{(s,s')\sim\caB}[\log{D_{\omega}(s,s')}] - \bbE_{(s,s')\sim\caD}[\log{(1-D_{\omega}(s,s')})]~,
       $$
       \ENDIF
       \STATE Sample $(s,a,r,s')\sim\caB$
       \STATE Update $\psi$ by $\caL_{\psi}^{\pi,h}$ (\eq{eq:final-loss})
       \ENDFOR
      
    \end{algorithmic}
\end{algorithm}

\section{Additional Related Works}
\label{ap:related}

\subsection{Transferable Imitation Learning}
Throughout this work, we've discussed several solutions in SOIL and works that utilize either inverse dynamics or state predictor in \se{sec:related} and summarized the difference among these works in \tb{tb:inverse-dynamics-comparison}. 
Moreover, in our paper, we pointed out the most important feature of DePO is that it allows transferable imitation learning to different action spaces or action dynamics. 
Before DePO, there are a few works have investigated transferable imitation learning. For instance, SOIL works can be essentially used for mimicking the expert state transition sampled with different action dynamics~\cite{liu2019state,jiang2020offline}, but they still learn an ad-hoc state-to-action policy which cannot be further transferred to other tasks; \citep{lee2021generalizable} proposed generalizable imitation learning for goal-directed tasks, but their generalized ability only limits in unseen states and goals in the same environment. 
Besides, \citep{srinivas2018universal} proposed an interesting reward transferring solution. Specifically, they imitate an action sequence using a gradient descent planner, which consists of an encoder module mapping the pixels into latent spaces. The encoder can be further used for constructing an obstacles-aware reward function by measuring the distance from the goal to the current observation. The authors conduct a series of experiments indicating that such an encoder and the corresponding reward function can be transferred to different environments, and even different robots to train different ad-hoc RL agents.

In comparison, the transferring ability of DePO comes from the pre-trained state planner model, which can be directly applied to different action spaces or dynamics, with little sampling cost on training the inverse dynamics model (or even no training at all if we have a ground truth inverse dynamics function serving as a control module).

\subsection{Relation to Hierarchical Reinforcement Learning}

Our work also relates a lot to Hierarchical Reinforcement Learning (HRL), since the role of the state planner behaves like a high-level policy and the inverse dynamics can be regarded as the low-level policy. 
In recent HRL works, a typical paradigm is to train the high-level policy using ground-truth rewards to predict the current goal for the low-level policy to achieve, while the low-level policy is trained by a handcrafted goal-matching reward function and provided the action to interact with the environment.
Notably, many HRL algorithms represent the goals within a learned embedding space (or know as option)~\cite{konidaris2007building,heess2016learning,kulkarni2016hierarchical,vezhnevets2017feudal,zhang2021hierarchical}, instead of the original state space; and these algorithms either keep a high-level action (option) for a fixed timesteps~\cite{nachum2018data,vezhnevets2017feudal} or learn a function to change the option~\cite{zhang2019dac,bacon2017option}. 
The most similar HRL work is \citep{nachum2018data}, whose high-level policy predicts in the raw form. However, there are lots of differences between their work and ours. First, they still train the high-level and low-level policies with separate reward functions and RL objectives, while our DePO optimizes both modules end-to-end; 
furthermore, their high-level policy holds a fixed goal for $c$ steps indicating the desire for the low-level policy to yield after $c$ steps, making it hard to be transferred;
finally, their main contribution lies in improving the learning efficiency on complicated tasks yet we provide a way for transferring the high-level state planner to different action spaces and dynamics. In \citep{heess2016learning}, they also try to transfer after learning the hierarchical modules; nevertheless, they pre-train and transfer the low-level controller which is fixed when training a different high-level controller. 
Like DePO, \cite{li2019sub} also optimized the hierarchical policy in an end-to-end way and derived the hierarchical policy gradient for optimizing the two-level policy jointly, but their high-level policy predicts a latent skill and fixes the decision for $c$ timesteps. In comparison, DePO obtains and fixes the low-level inverse dynamics (controllers) before optimizing the high-level state planner.

\setcounter{theorem}{0}
\setcounter{proposition}{0}

\section{Experiment}
\label{ap:exps}

\subsection{Experiment Settings}
\label{ap:exp-sets}

\subsubsection{Real-World Traffic Dataset}

NGSIM I-80 dataset includes three videos with a total length of 45 minutes recorded in a fixed area, from which 5596 driving trajectories of different vehicles can be obtained. The original data is equally split into three time intervals, each of which has the length of 15 minutes. We use the first split, among which we choose the first 1000 trajectories for our experiment. We choose 85\% of these trajectories as the training set and the remaining 15\% as the test set. In our experiment, the state space includes the position and velocity vectors of the ego vehicle and six neighbor vehicles, and the actions are vectors of acceleration and the change in steering angle.  

\begin{figure}[htb]
    \centering
    \includegraphics[width=0.4\textwidth]{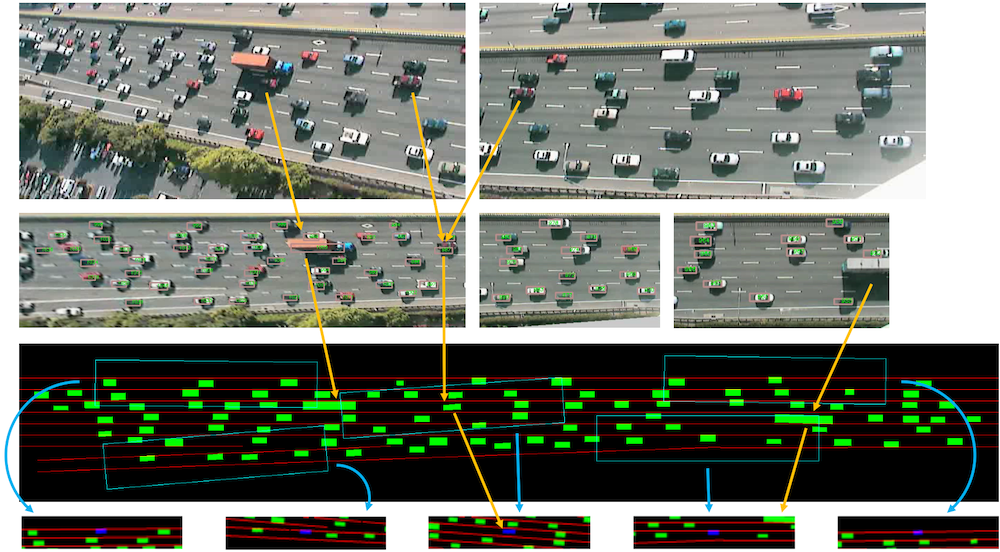}
    \caption{Visualization of NGSIM I-80 data set and its mapping on the simulator. This figure is borrowed from \citep{henaff2019model}.}
    \label{fig:ngsim-simulator}
\end{figure}

\subsubsection{Implementation Details}
For all experiments, we implement the value network as two-layer MLPs with 256 hidden units and the discriminator as 128 hidden units. For DePO and GAIfO-DP, the state predictor is a two-layer MLP with 256 hidden units and the inverse dynamics model is a four-layer MLP  with 512 hidden units. For GAIfO and BCO, the policy network is a two-layer MLP with 256 hidden units.
For fairness, we re-implement all the algorithms based on a Pytorch code framework\footnote{\url{https://github.com/Ericonaldo/ILSwiss}.} and adopt Soft Actor-Critic (SAC)~\citep{haarnoja2018soft} as the RL learning algorithm for GAIfO and DePO. All algorithms are trained with the same frequency and gradient steps.

For Mujoco benchmarks, we train an SAC agent to collect expert data, and take it for training the imitation learning agents. At training time we remove the terminal states and each episode will not end until 1000 steps. At testing time the terminal states are set for fair comparison. All algorithms are evaluated by deterministic policies. The codes are now public at \url{https://github.com/apexrl/DePO}.

For the NGSIM driving experiment, the original state contains the information of other cars, which is hard to predict. Therefore, we ignore it when predicting the state transition and the action of inverse dynamics. Formally, denote the state of all vehicles as $s$ and the state of the ego vehicle as $s_e$, the state planner predicts $s'_e$ from $s$:
$
s'_e \sim h(s'_e|s)~.
$
The inverse dynamics model predicts the controllable action $a$ based on the consecutive states of ego vehicle:
$
a \sim I(a|s_e,s'_e)~.
$
The discriminator is also constructed as differentiating the states of all vehicles $D(s,s)$. The codes are now public at \url{https://github.com/apexrl/DePO_NGSIM}.

During training, we randomly pick one car to be controlled by the policy at the beginning of every episode, and we replay the other cars by data. The episode ends when cars collide or successfully get through the road. To reduce the sampling time in the driving simulator, we implement parallel sampling using Python \textit{multiprocessing} library. In practice, we run 25 simulators to collect samples at the same time.

For training DePO on both Mujoco and NGSIM driving experiment, we normalize the Q value into an interval of $[0,1]$ for CDePG and keeps the Q value for DePG. Since CDePG (\eq{eq:weightedmle-sp}) can actually be treated as a weighted MLE objective as discussed in \ap{ap:ablation} and the supervised learning loss of the state planner (\eq{eq:sup-sp}) can also be implemented as an MLE objective, thereafter, in our implementation, we additionally allocate the maximum Q value (which is exactly 1 after normalization) as the weight for the supervised loss, and $\lambda_h$ also works for CDePG. In other words, we optimize the following objectives in these experiments:
\begin{equation}\label{eq:final-loss}
\begin{aligned}
    \min_{\phi}L^{I}~~\text{(if $I$ need to be learned)},\\
    \min_{\psi}\caL^{\pi,h} = \caL^{\pi}_{\text{DePG}} + \lambda_h (\caL^h + \caL^{\pi}_{\text{CDePG}}) ~,
\end{aligned}
\end{equation}

\subsubsection{Transferring Settings}
\label{ap:transfer-setting}

For transferring experiments on Mujoco, we try two kinds of dynamics setting:
i) \textbf{simple} transfer (results can be referred to \ap{ap:simple-transfer}), we invert the original action dynamics in each task; in other words, the original environment transition function takes the addictive inverse of the input action. For example, an action of -0.5 in the transferring experiments will lead to the same results as the action of 0.5 in the imitation experiments under the same state. ii) \textbf{complex} transfer (results can be referred to \se{sec:transfer}): we take an 80\% of the original gravity (exactly your proposal 1) with a complicated dynamics for the transferring experiment (different both action space and dynamics). Particularly, given the original action space dimension $m$ and dynamics $s'=f_s(a)$ on state $s$, the new action dimension and dynamics become $n=2m$ and $s'=f_s(h(a))$, where $h$ is constructed as:
\vspace{-7pt}
$$h= - \exp(a[0:n/2] + 1) + \exp(a[n/2:-1])) / 1.5~,\vspace{-7pt}$$
here $a[i:j]$ selects the $i$-th to $(j-1)$-th elements from the action vector $a$.
In other words, we transfer to a different gravity setting while doubling the action space and construct a more complicated action dynamics for agents to learn.

\subsubsection{Hyperparameters}
\label{ap:hyperpara}
We list the key hyperparameters of the best performance of DePO on each task in \tb{tb:hypers}. For each task, except the additional parameter -- state planner coefficient $\lambda_{h}$, the other hyperparameters are the same for all tested methods (if they need such parameters).

\begin{table*}[htbp]
\caption{Hyperparameters of DePO. Note that in our experiment, except the additional state planner coefficient $\lambda_{h}$, other hyperparameters are the same for all tested methods.}
\label{tb:hypers}
\vspace{1ex}
  \centering
  \resizebox{0.6\textwidth}{!}{
    \begin{tabular}{|l|c|c|c|c|c|c|}
    \hline
    Environments & Invert. & Hop. & Walk. & Half. & Human. & NGSIM. \\
    \hline
    Trajectory maximum length & \multicolumn{5}{c|}{1000} & 1500 \\
    \hline
    Optimizer & \multicolumn{6}{c|}{AdamOptimizer} \\
    \hline
    Discount factor $\gamma$ & \multicolumn{6}{c|}{0.99} \\
    \hline
    Replay buffer size & \multicolumn{5}{c|}{2e5} & 2e6 \\
    \hline
    Batch size & \multicolumn{5}{c|}{256} & 1024 \\
    \hline
    State planner coefficient $\lambda_{h}$ & \multicolumn{3}{c|}{0.1} & \multicolumn{2}{c|}{0.01} & 1.0 \\
    \hline
    $Q$ learning rate & \multicolumn{6}{c|}{3e-4} \\
    \hline
    $\pi$ learning rate & \multicolumn{6}{c|}{3e-4} \\
    \hline
    $D$ learning rate & \multicolumn{6}{c|}{3e-4} \\
    \hline
    $I$ learning rate & \multicolumn{6}{c|}{1e-4} \\
    \hline
    $I$ learning interval (epochs) & \multicolumn{6}{c|}{10} \\
    \hline
    Gradient penalty weight & \multicolumn{2}{c|}{4.0} & 8.0 & \multicolumn{1}{c|}{0.5} & 16.0 & 4.0\\
    \hline
    Reward scale & \multicolumn{6}{c|}{2.0} \\
    \hline
    \end{tabular} %
    }
\end{table*}

\subsection{Additional Experiment Results}
\label{ap:exps-res}

\subsection{Simple Transfer on Mujoco}
\label{ap:simple-transfer}
\begin{figure*}[t]
    \centering  
    \includegraphics[width=0.9\textwidth,height=2.95cm]{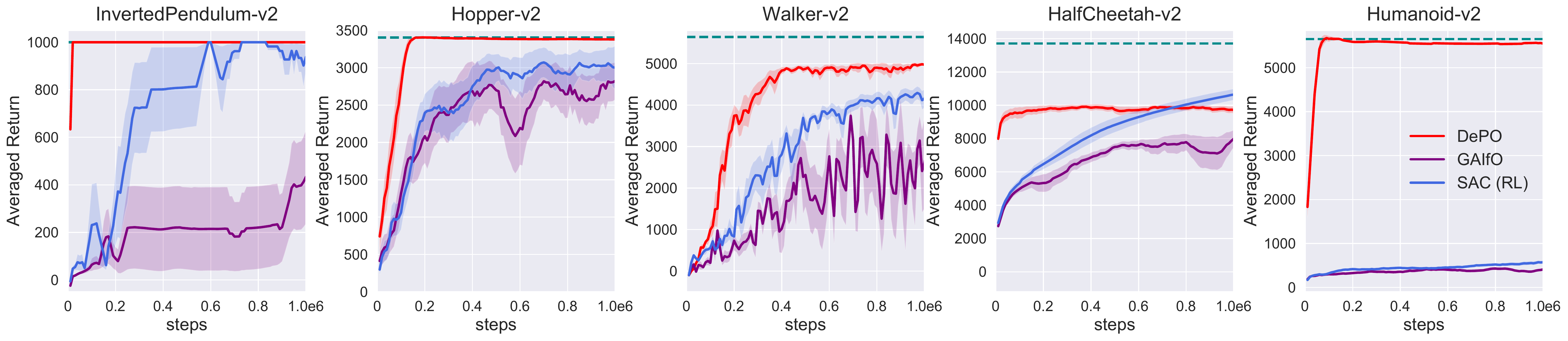}
    \vspace{-10pt}
    \caption{Transferring by pre-training on Mujoco tasks with inverted action dynamics within 1e6 interaction steps.}
    \vspace{-10pt}
    \label{fig:mujoco-transfer-naive}
\end{figure*}

In our early submission, we first try a naive transfer setting, i.e, inverted the action dynamics as stated in \ap{ap:transfer-setting}, to illustrate the advantage of DePO, shown in \fig{fig:mujoco-transfer-naive}. By comparing the results of two transferring scenarios, we can observe that the complex setting does not affect much to the performance of DePO, while the other baselines do not perform in complex transfer setting as well as they do in simple setting.

\subsubsection{Does DePO Reach Where it Predicts?}
\label{ap:reach-prediction}
\begin{figure*}[htbp]
    \centering
    \includegraphics[width=0.95\textwidth]{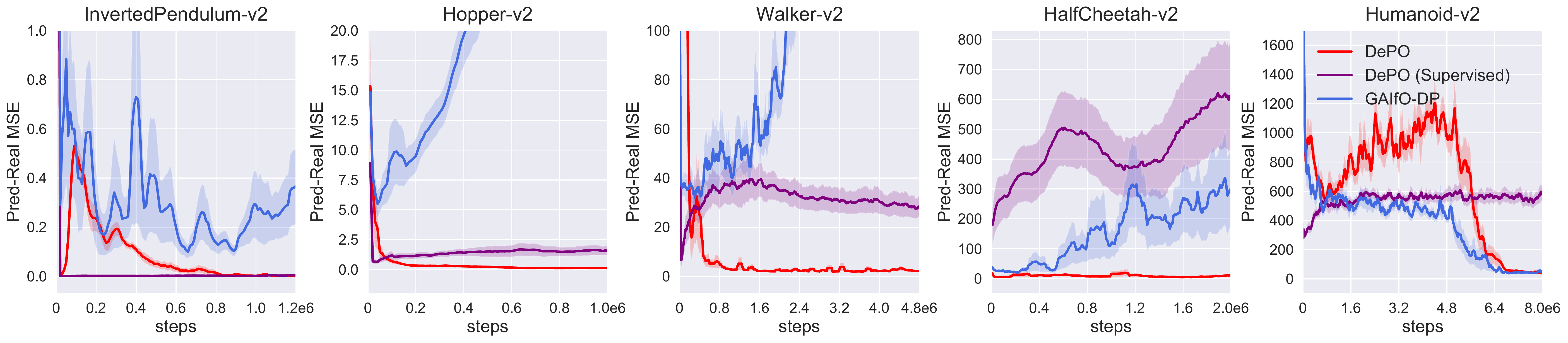}
    \vspace{-10pt}
    \caption{MSE curves of the one-step prediction of the state planner and the real state achieved in the environment for imitation experiments. The target state predictions are accurate and the differences are getting smaller during the training.}
    \label{fig:test-pred}
\end{figure*}

In this section, we aim to study whether the agent can reach where it plans to go. Therefore, we analyze the distance of the reaching states and the predicted consecutive states and draw the mean square error (MSE) along the imitation training procedure in \fig{fig:test-pred}. As shown in the figure, as the training goes, the averaged gap between the planned states and the states to achieve becomes smaller. On the contrary, the two variants, i.e., DePO (Supervised) and GAIfO-DP do not show any consistency. Particularly, DePO (Supervised) has little prediction error on two simple domains, but from the return curves, we believe that it stays at the initial state without any exploring; on the other tasks, the error does not even converge. Likewise, the prediction errors of GAIfO-DP diverge in most cases since there is no signal for the state planner to learn to predict the target state. The only exception is on Humanoid, where GAIfO-DP surprisingly converges as DePO does. We guess the reason may be the shallow layers of the policy (layers of state planner) learn an identical mapping, and the errors between two consecutive states are small. As for DePO, we further illustrate that the state planner does predict reasonable state transitions (\ap{ap:planner}), which is even accurate enough for multi-step planning. 

From our experience, the agent must have adequate exploration during the early learning stage, which may cause a large difference between the desired targets and the reached ones. If not, the agent tends to spin in a small local area, thus deteriorating the final performance. A piece of evidence showing such problems is the performance of DePO (Supervised). Except on HalfCheetah where the MSE explodes, the differences between the reaching states and the predictions on the other tasks are all limited. However, on Hopper and Walker, the imitation performance is bad since the prediction is only around the neighbor of the initialized state. 

\begin{figure*}[htbp]
    \centering
    \includegraphics[width=0.95\textwidth]{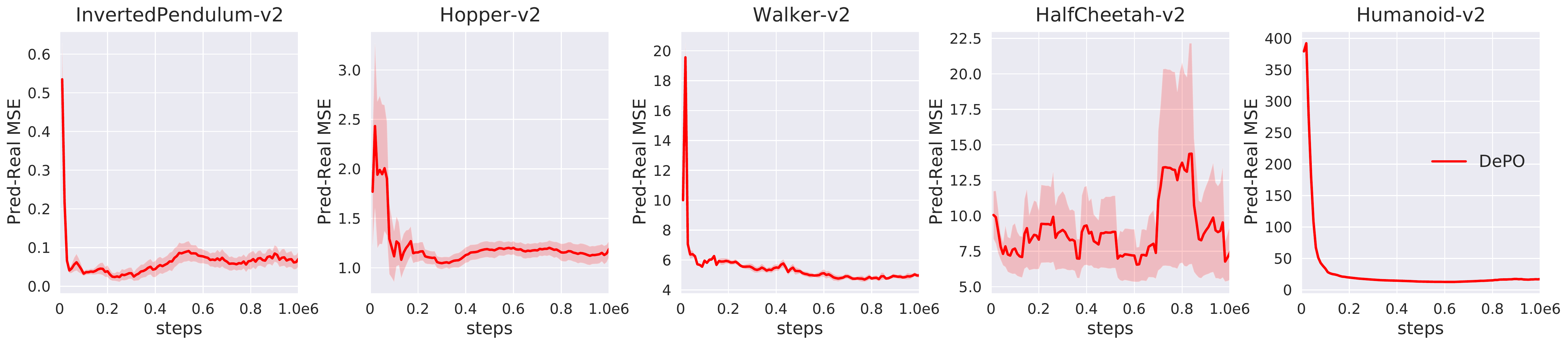}
    \vspace{-10pt}
    \caption{MSE curves of the one-step prediction of the state planner and the real state that the agent achieves in the environment in \textbf{complex} transfer experiments. The target state predictions are stably accurate along the whole training stage.}
    \label{fig:test-pred-transfer-complex}
\end{figure*}

\begin{figure*}[htbp]
    \centering
    \includegraphics[width=0.95\textwidth]{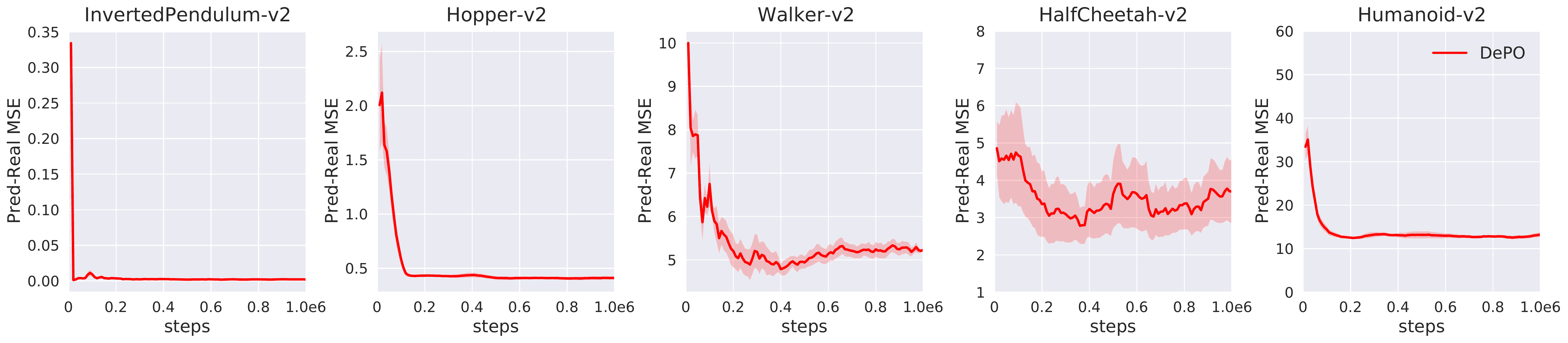}
    \vspace{-10pt}
    \caption{MSE curves of the one-step prediction of the state planner and the real state that the agent achieves in the environment in \textbf{simple} transfer experiments. The target state predictions are stably accurate along the whole training stage.}
    \label{fig:test-pred-transfer}
\end{figure*}

We also show the MSE curves during the transferring experiment in \fig{fig:test-pred-transfer-complex} and \fig{fig:test-pred-transfer}. As expected, the target state predictions of DePO are still stably accurate along the whole training stage. 

\newpage
\subsubsection{How Can the State Predictor be Used for Planning?}
\label{ap:planner}
\begin{figure*}[!h]
\begin{minipage}{\textwidth}
    \centering
    \includegraphics[width=0.845\textwidth]{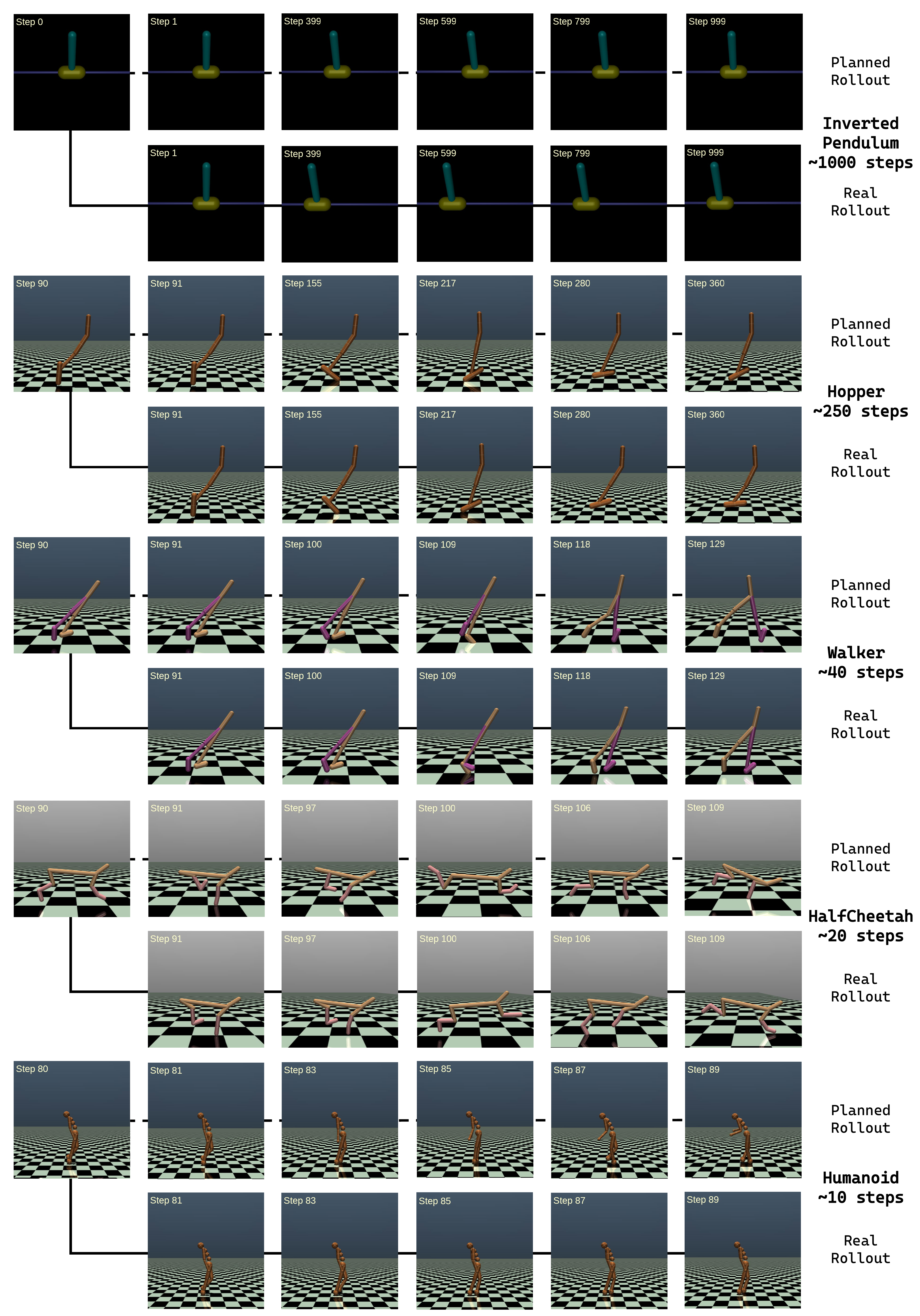}
    \vspace{-10pt}
    \caption{Imagined rollout by state planner and the real rollout that was achieved during the interaction with the environment on Mujoco tasks. We also provide a corresponding demo video in         \url{https://youtu.be/WahVjjvcYYM}.}
    \label{fig:mujoco-planner}
\end{minipage}
\end{figure*}

\begin{figure*}[!h]
\begin{minipage}{\textwidth}
    \centering
    \includegraphics[width=0.845\textwidth]{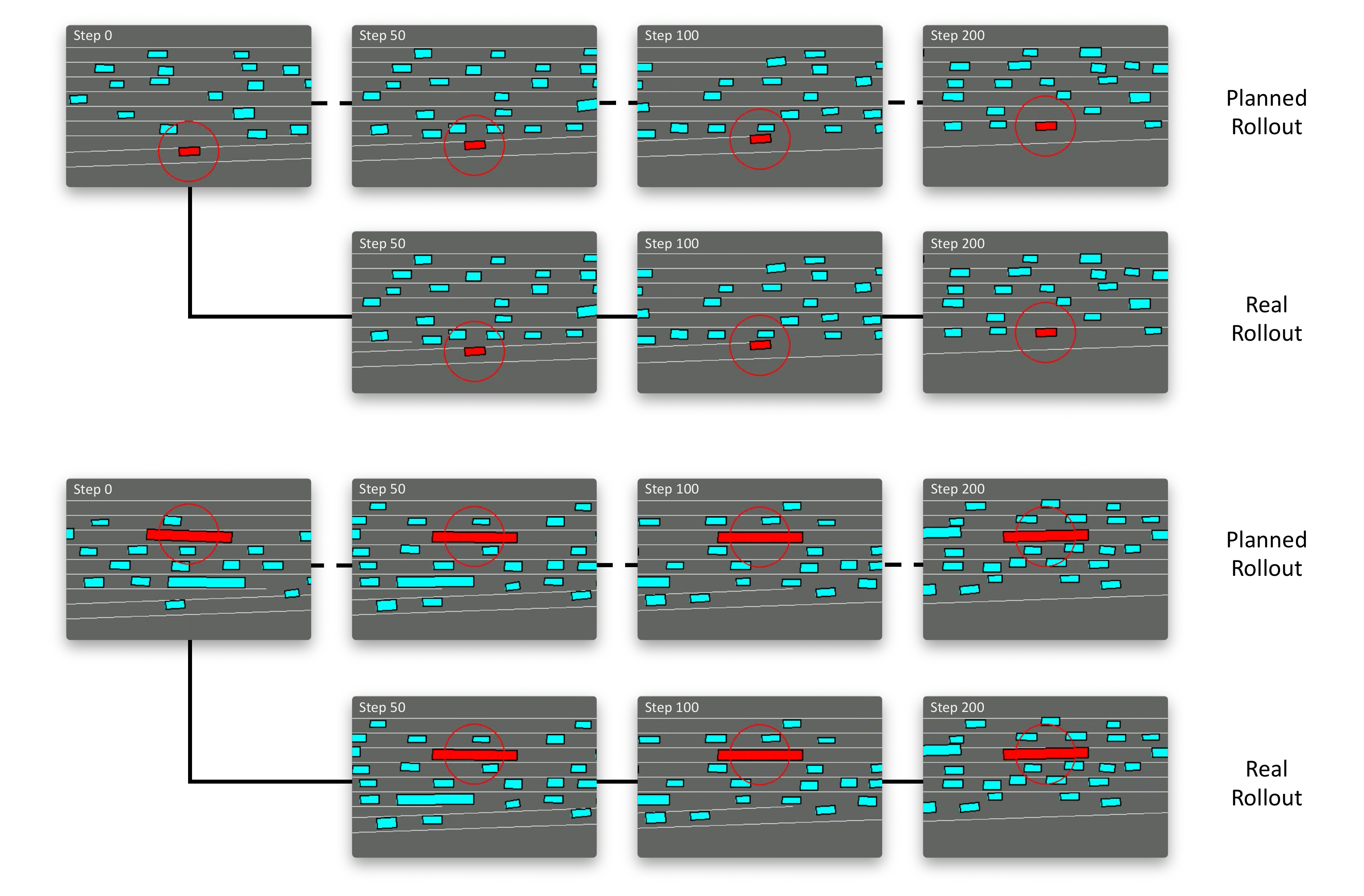}
    \vspace{-10pt}
    \caption{Imagined rollout by state planner and the real rollout that achieved during the interaction with the environment on NGSIM dataset.}
    \label{fig:ppuu-planner}
\end{minipage}
\end{figure*}

In our experiments, we find the state planner is rather accurate for predicting the next state with small compounding errors, and therefore we want to know how it can be used for multi-step planning without interacting with the environment. In this section, we compare the image of the imagined rollouts provided by the state planner and the real rollout that the agent achieves during its interactions with the environments. Note that the imagined rollouts are generated by consecutively predicting from predicted states. Formally, on a state $s$, $h_{\psi}$ predicts the next possible state $\hat{s}'$ to reach, then $\hat{s}'$ is further taken as an input to the predictor and predicts the two-step away state $\hat{s}''$. We repeat the cycle of prediction on prediction progressively until the imagined state makes no sense, i.e., the compounding error explodes. The results on Mujoco tasks are shown in \fig{fig:mujoco-planner} and NGSIM is in \fig{fig:ppuu-planner}. It is surprising that the planner can rollout at least tens or even hundreds of steps accurately.

\clearpage

\newpage

\subsection{Ablation Studies}
\label{ap:ablation}
\begin{figure}[htb]
\centering
\subfigure[Ablation study on $\lambda_h$.]{
\label{fig:ablation-lambda-h}
\begin{minipage}[b]{0.47\linewidth}
\label{fig:expert_heat}
\includegraphics[width=0.99\linewidth]{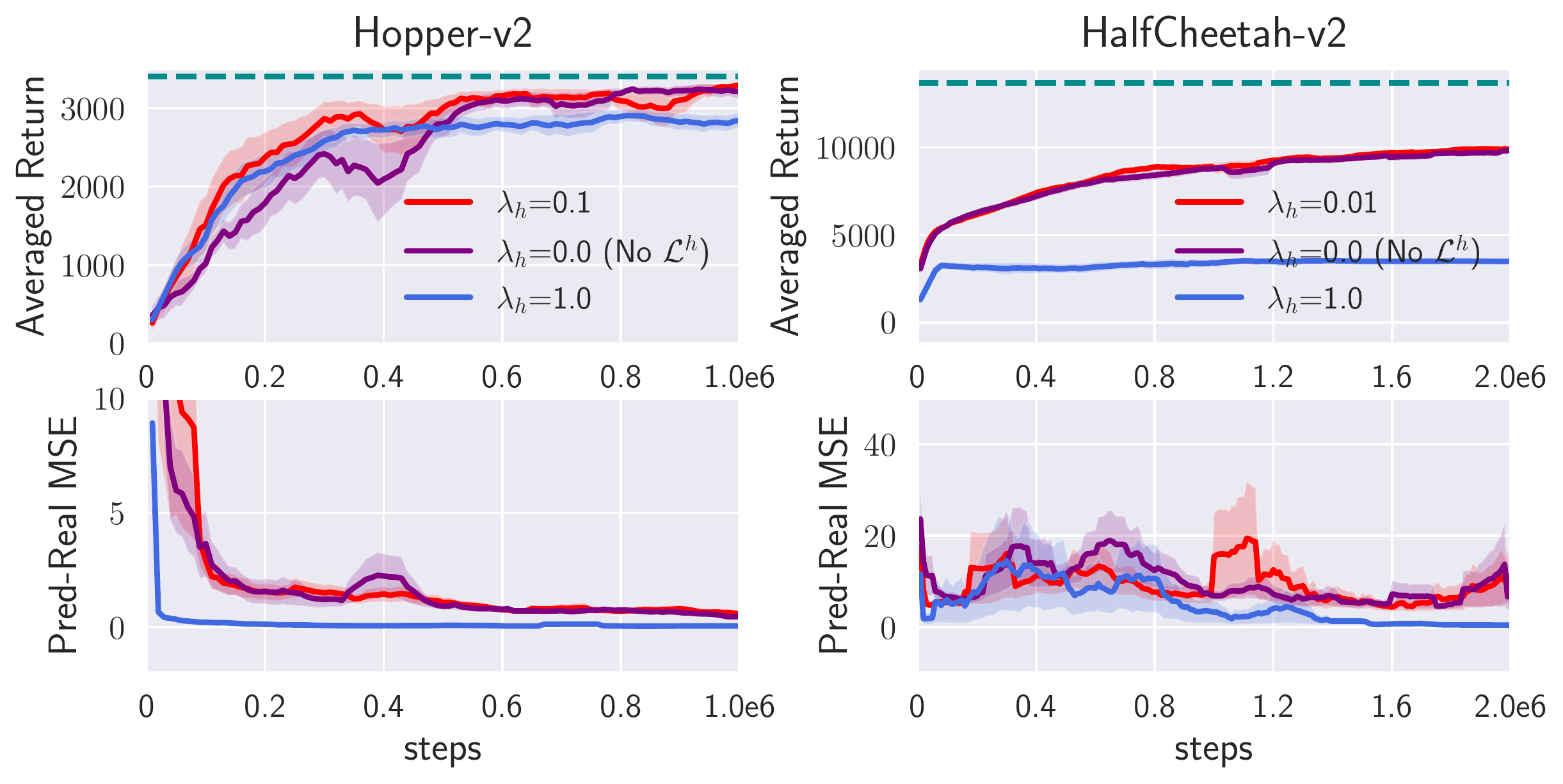}
\end{minipage}
}
\subfigure[Ablation study on CDePG.]{
\label{fig:ablation-cdepg}
\begin{minipage}[b]{0.47\linewidth}
\label{fig:ebil_heat}
\includegraphics[width=0.99\linewidth]{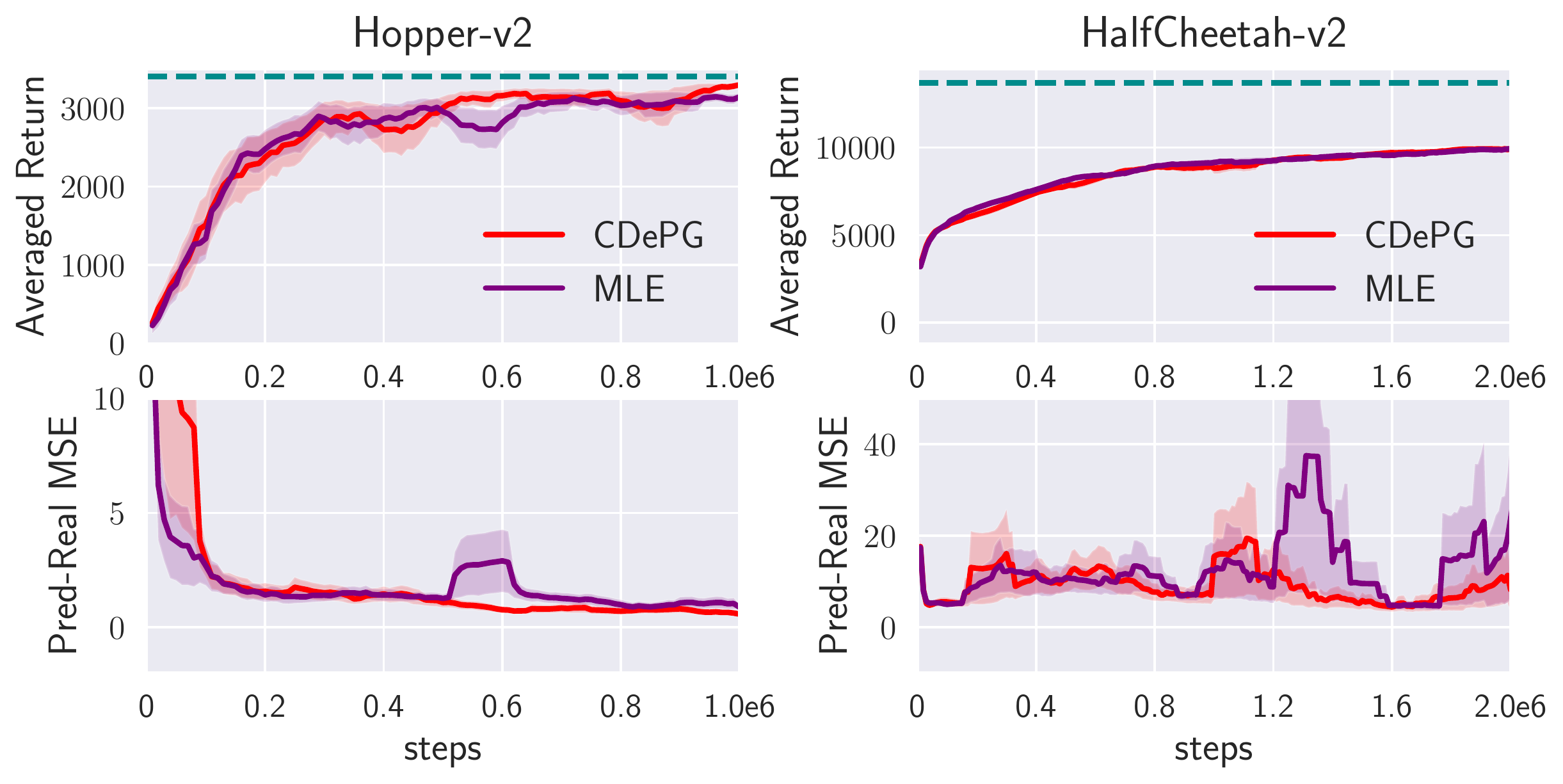}
\end{minipage}
}
\caption{Ablation studies.}
\end{figure}

\paragraph{Ablation on $\lambda_h$.}
For DePO, the only important hyperparameter is the supervised coefficient of state planner $\lambda_h$. In our method, we combine supervised learning with policy gradient to train the state planner. Therefore, we first investigate the effect of the supervised learning objective by setting $\lambda_h=0.0$ (i.e., removing $\caL^h$). The results are shown \fig{fig:ablation-lambda-h}, which indicates that the supervised learning actually does not affect much about the training procedure of DePO, and therefore can be removed. However, on some tasks like Hopper, augmenting expert supervision can lead to better learning efficiency.
In addition, a large $\lambda_h$ actually deteriorates the final performance, especially on harder tasks, while the prediction error can be even smaller. Therefore, we believe the deterioration comes from the poor exploration ability of the large $\lambda_h$, which asks high accuracy of the predictions on expert data even at the beginning of training. 

\paragraph{Ablation on CDePG.}

In \se{sec:depg}, we propose CDePG, which is used to alleviate the agnostic challenge of DePG by training the state planner with real experience sampled from the environment. From another point of view, updating CDePG (\eq{eq:weightedmle-sp}) is learning towards a weighted MLE objective (which is actually also the case for vanilla policy gradient), where the value function $Q(s, a)$ serving as the weight. To better understand how CDePG works, we conduct an ablation experiment on removing the weight $Q$, i.e., we are now optimizing an MLE objective (but also with coefficient $\lambda_h$ for better exploration) instead of CDePG. The results are shown in \fig{fig:ablation-lambda-h}, indicating that MLE actually achieves similar results to CDePG, although the MSE errors can be worse. From such a perspective, DePO can also be realized as regularizing an MLE constraint on DePG. In addition, without such regularization, DePO simply works bad with large MSE errors on all environments, due to the agnostic update of the state planner.

\subsection{Additional Results for NGSIM Experiments}
\label{ap:ppuu-exps}

\paragraph{Statistical evaluation metrics.}
We utilize \textit{Success Rate}, \textit{Mean Distance} and \textit{KL Divergence} as evaluation metrics for evaluating the final performances.
Specifically, \textit{Success Rate} is the percentage of driving across the entire area without crashing into other vehicles or driving off the road, \textit{Mean Distance} is the distance traveled before the episode ends, and \textit{KL Divergence} measures the position distribution distance between the expert and the agent. The results are presented in \tb{tb:ngsim-ppuu}. All metrics are taken average over the aforementioned three different action dynamics. Note that for each method, the performances are evaluated using the best model with the highest success rate.

\begin{table}[h]
\caption{Statistical performance on NGSIM I-80 driving task over 5 random seeds.}
\centering
\begin{tabular}{cccc}
\toprule
\multirow{2}{*}{Method} & Success & Mean & KL \\
& Rate (\%) & Distance (m) & Divergence \\
\midrule
GAIfO & 70.3 $\pm$ 2.5 & 180.0 $\pm$ 0.9 & 23.0 $\pm$ 10.2 \\
DePO/Co & 84.2 $\pm$ 5.2 & 192.4 $\pm$ 4.6 & 12.4 $\pm$ 7.1 \\
\midrule
Expert & 100.0 & 210.0 & 0.0 \\
\bottomrule
\label{tb:ngsim-ppuu}
\end{tabular}
\end{table}

\begin{figure*}[h]
    \centering
    \includegraphics[width=0.75\textwidth]{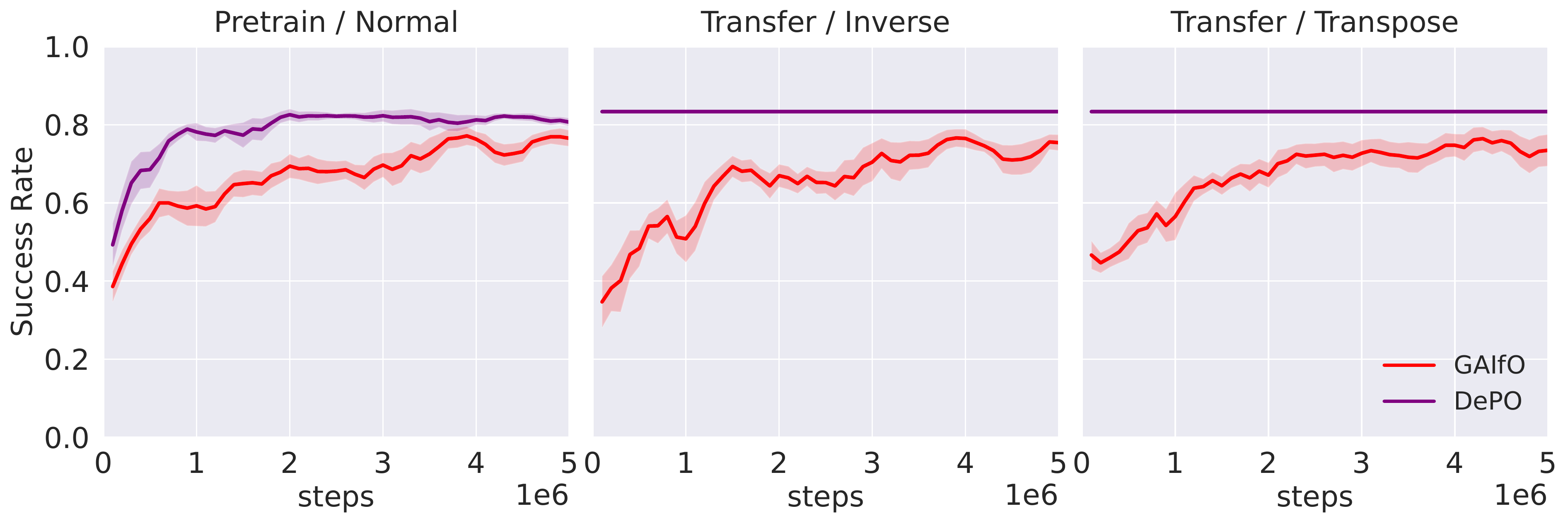}
    \caption{One-shot Transferring by pre-training on NGSIM dataset with different action dynamics.}
    \label{fig:ngsim-transfer}
\end{figure*}

\paragraph{One-shot transferring by pre-training.}
\label{ap:transfer:ppuu}
We illustrate the transfer ability of DePO with ground-truth inverse dynamics function on NGSIM I-80 datasets. Specifically, we train the high-level state-to-state state planner on the \textit{Normal} action dynamics, and then transferred to the \textit{Inverse} and \textit{Transpose} action dynamics. The results are shown in \fig{fig:ngsim-transfer}. As expected, with a known inverse dynamics function, the state planner can be directly deployed on different skills, which does not require additional sampling and training cost. However, the compared baseline method (GAIfO) needs to learn a state-to-action mapping policy for every agent separately.

\subsection{RL Training and Transferring}
\label{ap:transfer:rl}
\begin{figure*}[htbp]
    \centering
    \includegraphics[width=0.8\textwidth]{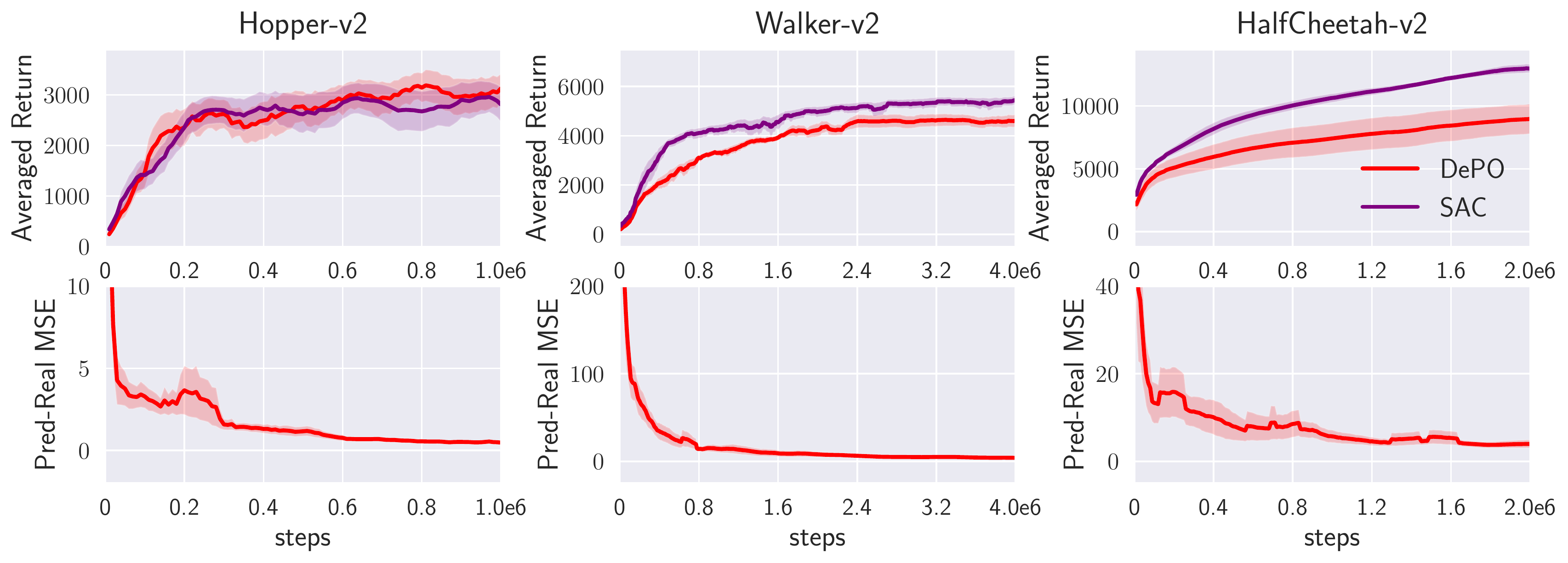}
    \vspace{-5pt}
    \caption{RL experiments on Mujoco tasks over 5 random seeds.}
    \label{fig:rl-train}
\end{figure*}

\begin{figure*}[htbp]
    \centering
    \includegraphics[width=0.8\textwidth]{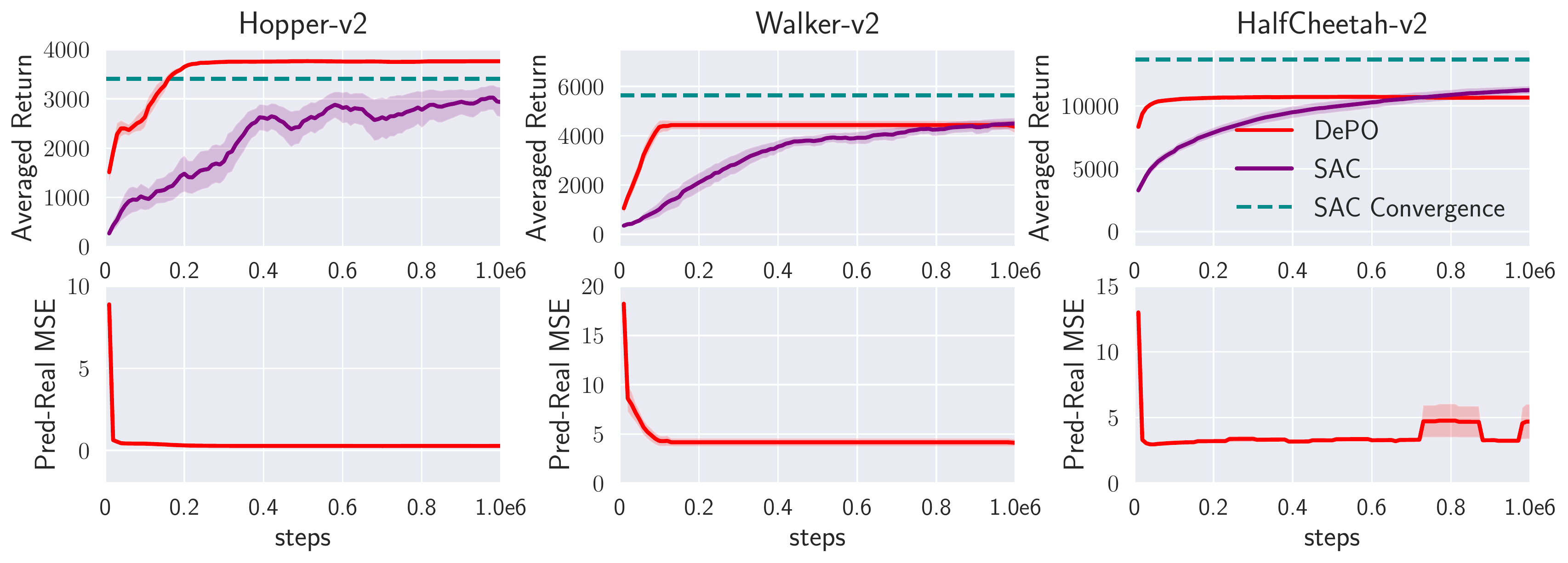}
    \vspace{-5pt}
    \caption{One-shot Transferring by pre-training RL agents on Mujoco tasks with inverted action dynamics over 5 random seeds.}
    \label{fig:rl-transfer}
\end{figure*}

\paragraph{RL training.}
We have mentioned that the key induction of DePG and CDePG does not limit in the literature of SOIL such that DePO can be applied to general RL tasks. In this section, we conduct RL experiments on Mujoco tasks comparing with SAC using the reward provided by the environments instead of learning from expert demonstrations. In this case, we do not have to keep a discriminator $D$ or optimize the supervised loss $\caL^h$. The learning curves are shown in \fig{fig:rl-train}, revealing the considerable performance of DePO. Although the learning efficiency and final performance may be lightly inferior to SAC, DePO can learn accurate high-level state planners that can be used for transferring (\fig{fig:rl-transfer}).

\paragraph{One-shot transferring by pre-training.}
From the bottom of \fig{fig:rl-train}, we know that DePO still learns accurate high-level state planners, and therefore we are able to transfer the learned planners to tasks that require different skills. We also test transferring on inverted action dynamics(\ap{ap:simple-transfer}), and the results are shown in \fig{fig:rl-transfer}. By pre-training a great state planner, DePO can achieve a stable performance on each task, and can even outperform the averaged convergence performance of SAC (on Hopper).

\section{Proofs}
\label{ap:proofs}

In our proofs we will work in finite state and action spaces $\caS$ and $\caA$ to avoid technical machinery out of the scope of this paper.

\begin{proposition}\label{approp:one-to-one} 
  Suppose $\Pi$ is the policy space and $\mathcal{P}$ is a valid set of state transition OMs such that $\mathcal{P}=\{\rho:\rho\geq 0\text{ and }\exists\pi\in\Pi, \text{s.t.}\; \rho(s, s')=\rho_0(s)\int_{a} \pi(a|s)\caT(s'|s, a)\dif a+\int_{s'', a} \pi(a|s)\caT(s'|s, a)\rho(s'', s)\dif s''\dif a\}$, then a policy $\pi\in\Pi$ corresponds to one state transition OM $\rho_{\pi}\in\mathcal{P}$. However, under the action-redundant assumption about the dynamics $\caT$, a state transition OM $\rho\in\mathcal{P}$ can correspond to more than one policy in $\Pi$.
\end{proposition}

\begin{proof}

We first provide the proof for the one-to-one correspondence between marginal distribution $\sum_{a}\pi(a|s)\mathcal{T}(s'|s,a)$ and state transition OM $\rho(s, s')\in\mathcal{P}$.

For a given policy $\pi$, by definition of state transition OM, we have

\begin{equation}
\begin{aligned}
\rho_{\pi}(s, s')&=\sum_{a}\mathcal{T}(s'|s, a)\rho_{\pi}(s,a)\\
&=\sum_{a}\pi(a|s)\mathcal{T}(s'|s, a)\sum_{t=0}^{\infty}\gamma^t P(s_t=s|\pi)~.
\end{aligned}
\label{eq:rho-s-s'-new}
\end{equation}

For all $t$ greater than or equal to 1, we have

\begin{equation}
\label{eq:recursive-prob-new}
P(s_t=s|\pi)=\sum_{s''} P(s_{t-1}=s'', s_{t}=s|\pi)~.
\end{equation}

Take \eq{eq:recursive-prob-new} into \eq{eq:rho-s-s'-new}, we have

\begin{equation}
\begin{aligned}
\rho_{\pi}(s, s'|\pi) &= P(s_0=s, s_{1}=s'|\pi)+\sum_{a}\pi(a|s)\mathcal{T}(s'|s, a)\sum_{t=1}^{\infty}\gamma^t\sum_{s''} P(s_{t-1}=s'', s_{t}=s|\pi)\\
&= P(s_0=s, s_1=s'|\pi)+\gamma\sum_{a}\pi(a|s)\mathcal{T}(s'|s, a)\sum_{s''}\sum_{t=0}^{\infty}\gamma^{t} P(s_{t}=s'', s_{t+1}=s|\pi) \\
&= P(s_0=s, s_1=s'|\pi)+\gamma\sum_{a}\pi(a|s)\mathcal{T}(s'|s, a)\sum_{s''}\rho_{\pi}(s'', s) \\
&= \rho_0(s)\sum_{a}\pi(a|s)\mathcal{T}(s'|s, a)+\gamma\sum_{s'', a}\pi(a|s)\mathcal{T}(s'|s, a)\rho_{\pi}(s'', s)
\end{aligned}
\label{eq:recursive-rho-new}
\end{equation}



Consider the following equation of variable $\rho$:

\begin{equation}
    \rho(s, s') = \rho_0(s)\sum_{a}\pi(a|s)\mathcal{T}(s'|s, a)+\gamma\sum_{s'', a}\pi(a|s)\mathcal{T}(s'|s, a)\rho(s'', s)~.
\label{eq:recursive-pi-new}
\end{equation}

According to \eq{eq:recursive-rho-new}, $\rho_{\pi}$ is a solution of \eq{eq:recursive-pi-new}. Now we proceed to prove $\rho_{\pi}$ as the unique solution of \eq{eq:recursive-pi-new}.





Define the matrix

\begin{equation*}
A_{(ss', s''s)}\triangleq \left\{
    \begin{array}{ll}
    1-\gamma \sum_{a}\pi(a|s)\mathcal{T}(s'|s, a) & \text{if} \; (s, s') = (s'', s) \\  
    -\gamma \sum_{a}\pi(a|s)\mathcal{T}(s'|s, a) & \text{otherwise}~.\\ 
    \end{array}  
\right.
\end{equation*}

Note that $A$ is a two-dimensional matrix indexed by state transition pairs. Also define the vector

\begin{equation*}
b_{s, s'}\triangleq \rho_0(s)\sum_{a}\pi(a|s)\mathcal{T}(s'|s, a)~.
\end{equation*}

We can rewrite \eq{eq:recursive-pi-new} equivalently as

\begin{equation}
A\rho=b~.
\end{equation}

Since $\sum_{s', a}\pi(a|s)\mathcal{T}(s'|s, a)=1$ and $\gamma<1$, for all $(s'', s)$, we have

\begin{equation*}
\begin{aligned}
&\quad\sum_{s, s'}\gamma \sum_{a}\pi(a|s)\mathcal{T}(s'|s, a)=\gamma < 1 \\
&\Rightarrow 1 - \gamma\sum_{a}\pi(a|s'')\mathcal{T}(s|s'', a) > \sum_{(s, s')\neq(s'', s)}\gamma \sum_{a}\pi(a|s)\mathcal{T}(s'|s, a)\\
&\Rightarrow \left| A_{(s''s, s''s)} \right|\geq\sum_{(s, s')\neq(s'', s)}\left| A_{(ss', s''s)} \right|~.
\end{aligned}
\end{equation*}

Therefore, we have proven $A$ as column-wise strictly diagonally dominant, which implies that $A$ is non-singular, so \eq{eq:recursive-pi-new} has at most one solution. Since for all $\rho$ in $\mathcal{P}$, it must satisfy the constraint \eq{eq:recursive-pi-new}, which means that for any marginal distribution $\sum_{a}\pi(a|s)\mathcal{T}(s'|s, a)$, there is only one corresponding $\rho$ in $\mathcal{P}$.

Now, we proceed to prove that for every $\rho$ in $\mathcal{P}$, there is only one corresponding marginal distribution $\sum_{a}\pi(a|s)\mathcal{T}(s'|s, a)$ such that $\rho_{\pi}=\rho$. By definition of $\mathcal{P}$, $\rho$ is the solution of \eq{eq:recursive-pi-new} for some policy $\pi$. By rewriting \eq{eq:recursive-pi-new}, the marginal distribution can be written in the form of a function expression of $\rho$ as

\begin{equation}
    \sum_a \pi(a|s)\mathcal{T}(s'|s, a)=\frac{\rho(s, s')}{\rho_{0}(s)+\gamma\sum_{s''}\rho(s'', s)}~.
\end{equation}

This means every $\rho\in\mathcal{P}$ only corresponds to one marginal distribution $\sum_a\pi(a|s)\mathcal{T}(s'|s,a)$. As we discussed before, $\rho$ is the state transition OM of $\pi$, i.e., $\rho=\rho_{\pi}$.

By establishing the one-to-one correspondence between the marginal distribution $\sum_a\pi(a|s)\mathcal{T}(s'|s,a)$ and state transition OM $\rho\in\mathcal{P}$, we can alternatively study the correspondence between the marginal distribution and policy. Obviously, one policy can only correspond to one marginal distribution. We now prove that if the dynamics $\mathcal{T}$ has redundant actions, one marginal distribution can correspond to more than one policy in $\pi$. 

We prove the statement by counterexample construction. If the dynamics $\mathcal{T}$ has redundant actions, there exist $s_m \in \mathcal{S}$, $a_n\in \mathcal{A}$ and distribution $p$ defined on $\mathcal{A}\setminus\{a_n\}$ such that $\sum_{a\in\mathcal{A}\setminus \{a_n\}}p(a)\mathcal{T}(s'|s_m, a)=\mathcal{T}(s'|s_m, a_n)$. Consider two policy $\pi_0$ and $\pi_1$ such that

\begin{equation}
    \left\{
        \begin{aligned}
            &\,\pi_0(a|s)=\pi_1(a|s) &\text{if}\;s\neq s_m\\
            &\,\pi_0(a_n|s_m) = 1 &\\
            &\,\pi_0(a|s_m) = 0 &\text{if}\;a\neq a_n\\
            &\,\pi_1(a_n|s_m) = 0 &\\
            &\,\pi_1(a|s_m) = p(a) &\text{if}\;a\neq a_n~.
        \end{aligned}
    \right.
    \label{eq:two-policy}
\end{equation}

From \eq{eq:two-policy}, we know that $\pi_0$ and $\pi_1$ are two different policies. However, they share the same marginal distribution $\sum_a\pi(a|s)\mathcal{T}(s'|s,a)$. To justify this, we first consider the case when $s$ equals to $s_m$, where we have

\begin{equation}
\begin{aligned}
    \sum_a\pi_0(a|s_m)\mathcal{T}(s'|s_m, a)&=\pi_0(a_n|s_m)\mathcal{T}(s'|s_m, a_n)+\sum_{a\in\mathcal{A}\setminus\{a_n\}}\pi_0(a|s_m)\mathcal{T}(s'|s_m, a)\\
    &=\mathcal{T}(s'|s_m, a_n)\\
    &=\sum_{a\in\mathcal{A}\setminus \{a_n\}}\pi_1(a|s_m)\mathcal{T}(s'|s_m, a)\\
    &=\sum_{a\in\mathcal{A}\setminus \{a_n\}}\pi_1(a|s_m)\mathcal{T}(s'|s_m, a)+\pi_1(a_n|s_m)\mathcal{T}(s'|s_m, a_n)\\
    &=\sum_a\pi_1(a|s_m)\mathcal{T}(s'|s_m, a)~.
\end{aligned}
\label{eq:equal-marginal}
\end{equation}

When $s$ does not equal to $s_m$, the equality holds trivially, since the action selection probability of $\pi_0$ and $\pi_1$ defined on these states are exactly the same. Thus, one marginal distribution can correspond to more than one policy in $\pi$ when there are redundant actions.

\end{proof}

\begin{proposition}
  Suppose the state transition predictor $h_{\Omega}$ is defined as in \eq{eq:state-transition} and $\Gamma=\{h_{\Omega}: \Omega \in\Lambda\}$ is a valid set of the state transition predictors, $\mathcal{P}$ is a valid set of the state-transition OMs defined as in \prop{prop:one-to-one}, then
  a state transition predictor $h_{\Omega}\in\Gamma$ corresponds to one state transition OM $\rho_{\Omega}\in\mathcal{P}$; and a state transition OM $\rho\in\mathcal{P}$ only corresponds to one hyper-policy state transition predictor such that $h_{\rho} = \rho(s,s')/\int_{s'}\rho(s,s')\dif s'$.
\end{proposition}

\begin{proof}

During the proof of \prop{prop:one-to-one}, we have an intermediate result that there is one-to-one correspondence between the marginal distribution $\sum_a\pi(a|s)\mathcal{T}(s'|s,a)$ and state transition OM $\rho\in\mathcal{P}$. Since the definition of state transition predictor is exactly $h_{\Omega}(s'|s)=\sum_a\pi(a|s)\mathcal{T}(s'|s,a)$ ($\forall \pi\in\Omega$), the one-to-one correspondence naturally holds between state transition predictor $h(s'|s)$ and state transition OM $\rho\in\mathcal{P}$.

\end{proof}

\begin{theorem}
    Let \caA(s) be an action set that for all $a$ in $\caA(s)$, a deterministic transition function leads to the same state $s'=\caT(s,a)$. If there exists an optimal policy $\pi^*$ and a state $\hat{s}$ such that $\pi^*(\cdot|\hat{s})$ is a distribution over $\caA(\hat{s})$, then we can replace $\pi^*(\cdot|\hat{s})$ with any distributions over $\caA(\hat{s})$ which does not affect the optimality.
\end{theorem}

\begin{proof}

We denote the policy after replacing $\pi^*(\cdot|\hat{s})$ with a distribution $p$ over $\mathcal{A}(\hat{s})$ as $\pi$, and we have

\begin{equation}
    \pi(a|s)=\left\{
        \begin{aligned}
        &\pi^*(a|s),  & s\neq\hat{s}, \\
        &p(a), & s=\hat{s}.
        \end{aligned}
    \right.
\end{equation}

The Bellman equation of $V^{\pi^*}$ can be written as

\begin{equation}
    V^{\pi^*}(s)=\bbE_{a\sim\pi^*(\cdot|s), s'\sim\mathcal{T}(\cdot|s, a)}[r(s, a)+V^{\pi^*}(s')]~.
\end{equation}

Since $\pi$ and $\pi^*$ only differ at state $\hat{s}$, for all $s\neq\hat{s}$, we have

\begin{align}
    V^{\pi^*}(s)&=\bbE_{a\sim\pi^*(\cdot|s), s'\sim\mathcal{T}(\cdot|s, a)}[r(s, a)+V^{\pi^*}(s')]\notag\\
    &=\bbE_{a\sim\pi(\cdot|s), s'\sim\mathcal{T}(\cdot|s, a)}[r(s, a)+V^{\pi^*}(s')]~.
    \label{eq:nohats-bellman}
\end{align}

Notice that for all $a$ in $\caA(\hat{s})$, the deterministic transition function lead to the same state $\hat{s}'=\caT(\hat{s},a)$. Since our reward function is state-only and defined on $(s, s')$, the reward $r(\hat{s}, a)=r(\hat{s},\hat{s}')$ must be the same for all $a$ in $\mathcal{A}(\hat{s})$. Therefore, we have

\begin{align}
    V^{\pi^*}(\hat{s})&=\bbE_{a\sim\pi^*(\cdot|\hat{s})}[r(\hat{s}, a)]+V^{\pi^*}(\hat{s}')\notag\\
    &=r(\hat{s}, \hat{s}')+V^{\pi^*}(\hat{s}')\notag\\
    &=\bbE_{a\sim p(\cdot)}[r(\hat{s}, a)]+V^{\pi^*}(s')~,
    \label{eq:hats-bellman}
\end{align}

which means \eq{eq:nohats-bellman} also holds for $\hat{s}$. Thus we reach the result that

\begin{equation}
    \forall s,\;V^{\pi^*}(s)=\bbE_{a\sim\pi(\cdot|s), s'\sim\mathcal{T}(\cdot|s, a)}[r(s, a)+V^{\pi^*}(s')]~.
\end{equation}

According to policy evaluation theorem, we know that $V^{\pi^*}$ is also the value function of policy $\pi$. Since $V^\pi$ is the optimal value function, we complete the proof that $\pi$ is also an optimal policy.

\end{proof}

\begin{theorem}[Error Bound of DePO]\label{theorem:compounding-error-dpo}
    Consider a deterministic environment whose transition function $\caT(s,a)$ is deterministic and $L$-Lipschitz. Assume the ground-truth state transition $h_{\Omega_E}(s)$ is deterministic, and for each policy $\pi\in\Pi$, its inverse dynamics $I_{\pi}$ is also deterministic and $C$-Lipschitz. Then for any state $s$, the distance between the desired state $s'_E$ and reaching state $s'$ sampled by the decoupled policy is bounded by:
\begin{equation}
    \|s'-s'_E\| \leq LC \| h_{\Omega_E}(s) - h_{\psi}(s) \| + L\|I_{\tilde{\pi}}(s,\hat{s}') - I_{\phi}(s,\hat{s}')\|~,
\end{equation}
where $\tilde{\pi}$ is a sampling policy that covers the state transition support of the expert hyper-policy and $\hat{s}' = h_{\psi}(s)$ is the predicted consecutive state.
\end{theorem}
\begin{proof}
Given a state $s$, the expert takes a step in a deterministic environment and get $s'$. We assume that the expert $\Omega_E$ can use any feasible policy $\tilde{\pi}$ that covers the support of $\Omega_E$ to reach $s$:
\begin{equation}
    s'_E = \caT(s, I_{\tilde{\pi}}(s, h_{\Omega}(s)))
\end{equation}
Similarly, using decoupled policy, the agent predict $\hat{s}'=h_{\psi}(s)$ and infer an executing action by an inverse dynamics model $a=I_{\phi}(s,s')$, which is learned from the sampling policy $\tilde{\pi}$. Denote the reaching state of the agent as $s'$:
\begin{equation}
    s' = \caT(s, I_{\phi}(s, h_{\psi}(s)))
\end{equation}
Therefore, the distance between $s'$ and $s'_E$ is:
\begin{equation*}
\begin{aligned}
    \|s'-s'_E\| = \|\caT(s, I_{\tilde{\pi}}(s, h_{\Omega}(s))) - \caT(s, I_{\phi}(s, h_{\psi}(s)))\|
\end{aligned}
\end{equation*}
Let's consider the deterministic transition on $s$ is a function of $a$ such that $s'=\caT^s(a)$, then we continue the deviation:
\begin{equation*}
\begin{aligned}
    \|s'-s'_E\| &\leq \|\caT^s(I_{\tilde{\pi}}(s, h_{\Omega}(s))) - \caT^s(I_{\phi}(s, h_{\psi}(s)))\|\\
    &\leq L\|I_{\tilde{\pi}}(s, h_{\Omega}(s))) - I_{\phi}(s, h_{\psi}(s))\|\\
    &\leq L\|I_{\tilde{\pi}}(s, h_{\Omega}(s))) - I_{\tilde{\pi}}(s, h_{\psi}(s))) + I_{\tilde{\pi}}(s, h_{\psi}(s))) - I_{\phi}(s, h_{\psi}(s))\|
\end{aligned}
\end{equation*}
Similarly we also take the inverse transition on $s$ is a function of $s'$ such that $a=I^s(s')$, then we have that:
\begin{equation}
\begin{aligned}
    \|s'-s'_E\| &\leq L\|I_{\tilde{\pi}}^s(h_{\Omega}(s))) - I_{\tilde{\pi}}^s(h_{\psi}(s))) \\
    &~~~~+ I_{\tilde{\pi}}^s(h_{\psi}(s))) - I_{\phi}^s(h_{\psi}(s))\|\\
    &\leq L\|I_{\tilde{\pi}}^s(h_{\Omega}(s))) - I_{\tilde{\pi}}^s(h_{\psi}(s)))\| + L\|I_{\tilde{\pi}}^s(h_{\psi}(s))) - I_{\phi}^s(h_{\psi}(s))\|\\
    &\leq LC\|h_{\Omega}(s)) - h_{\psi}(s))\| + L\|I_{\tilde{\pi}}^s(\hat{s}') - I_{\phi}^s(\hat{s}')\|~.
\end{aligned}
\end{equation}
\end{proof}

\begin{theorem}[Error Bound of BCO]\label{theorem:compounding-error-bco}
    Consider a deterministic environment whose transition function $\caT(s,a)$ is deterministic and $L$-Lipschitz, and a parameterized policy $\pi_{\psi}(a|s)$ that learns from the label provided by a parameterized inverse dynamics model $I_{\phi}$. Then for any state $s$, the distance between the desired state $s'_E$ and reaching state $s'$ sampled by a state-to-action policy as BCO~\citep{torabi2018behavioral} is bounded by:
\begin{equation}
\begin{aligned}
    \|s'-s'_E\| &\leq L\left \|\pi_{\psi}(a|s) - \int_{s'^{*}}p_{\piE}(s'^{*}|s)I_{\phi}(a|s, s'^{*})\dif s'^{*})\right \| \\
    &\quad + L\left \|\int_{s'^{*}}p_{\piE}(s'^{*}|s)I_{\tilde{\pi}}(a|s, s'^{*})) - p_{\piE}(s'^{*}|s)I_{\phi}(a|s, s'^{*})\dif s'^{*}\right \|~,
\end{aligned}
\end{equation}
where $\tilde{\pi}\in\omega_E$ is a policy instance of the expert hyper-policy $\omega_E$ such that $\caT(s, \tilde{\pi}(s)) = s'_E$.
\end{theorem}
\begin{proof}
\begin{equation}
\begin{aligned}
    \|s'-s'_E\| &= \|\caT(s, \pi_{\psi}(s)) - \caT(s, \tilde{\pi}(s))\|\\
    &= \|\caT^s(\pi_{\psi}(s)) - \caT^s(\tilde{\pi}(s))\|\\
    &\leq L\|\tilde{\pi}(a|s) - \pi_{\psi}(a|s)\|\\
    &= L\left\|\pi_{\psi}(a|s) - \int_{s'^{*}}p_{\piE}(s'^{*}|s)I_{\phi}(a|s, s'^{*})\dif s'^{*} \right.\\
    &\quad +\left.\int_{s'^{*}}p_{\piE}(s'^{*}|s)I_{\phi}(a|s, s'^{*})\dif s'^{*} - \int_{s'^{*}}p_{\piE}(s'^{*}|s)I_{\tilde{\pi}}(a|s, s'^{*}))\dif s'^{*}\right \|\\
    &\leq L\left \|\pi_{\psi}(a|s) - \int_{s'^{*}}p_{\piE}(s'^{*}|s)I_{\phi}(a|s, s'^{*})\dif s'^{*})\right \| \\
    &\quad+ L\left \|\int_{s'^{*}}p_{\piE}(s'^{*}|s)I_{\tilde{\pi}}(a|s, s'^{*})) - p_{\piE}(s'^{*}|s)I_{\phi}(a|s, s'^{*})\dif s'^{*}\right \|
\end{aligned}
\end{equation}
\end{proof}
An intuitive explanation for the bound is that BCO~\citep{torabi2018behavioral} first seeks to recover a policy that shares the same hyper-policy with $\piE$ via learning an inverse dynamics model and then try to conduct behavior cloning. Therefore the errors come from the reconstruction error of $\tilde{\pi}$ using $I_{\phi}$ (the second term) and the fitting error of behavior cloning (the first term).

By comparing \theo{theorem:compounding-error-dpo} and \theo{theorem:compounding-error-bco}, it is observed that for reaching each state, BCO requires a good inverse dynamics model over the state space to construct $\tilde{\pi}$ and then conduct imitation learning to $\tilde{\pi}$, while DePO only requires to learn a good inverse dynamics model on the predicted state and directly construct $\tilde{\pi}$ without the second behavior cloning step. 